\documentclass[aoas]{imsart}
\usepackage{bm}
\usepackage{xcolor}
\usepackage{bbold}
\usepackage{multirow}
\usepackage{booktabs} 
\usepackage{float}
\usepackage[para,online,flushleft]{threeparttable}

\usepackage{lipsum}

\newcommand\blfootnote[1]{%
  \begingroup
  \renewcommand\thefootnote{}\footnote{#1}%
  \addtocounter{footnote}{-1}%
  \endgroup
}

\RequirePackage{amsthm,amsmath,amsfonts,amssymb}
\RequirePackage[authoryear]{natbib}
\RequirePackage[colorlinks,citecolor=blue,urlcolor=blue]{hyperref}
\RequirePackage{graphicx}

\startlocaldefs
\theoremstyle{plain}

\newtheorem{theorem}{Theorem}[section]

\theoremstyle{remark}

\newtheorem{assumption}{Assumption}

\usepackage{algorithm}
\usepackage{algpseudocode}

\usepackage{algpseudocode}
\newcommand\Algphase[1]{ \vspace*{-.5\baselineskip}\Statex\hspace*{\dimexpr-\algorithmicindent-2pt\relax}\rule{14.2cm}{0.4pt}%
\Statex\hspace*{-\algorithmicindent}\textbf{#1}%
\vspace*{-.5\baselineskip}\Statex\hspace*{\dimexpr-\algorithmicindent-1pt\relax}\rule{14.2cm}{0.4pt}}%

\endlocaldefs

\begin{document} 

\begin{frontmatter}
\title{CeCNN: Copula-enhanced convolutional neural networks in joint prediction of refraction error and axial length based on ultra-widefield fundus images}
\runtitle{Copula enhanced CNNs}

\begin{aug}
\author[A]{\fnms{Chong}~\snm{Zhong$^{\dag}$}},
\author[B]{\fnms{Yang}~\snm{Li$^{\dag}$}}, 
\author[D]{\fnms{Danjuan}~\snm{Yang}},
\author[D]{\fnms{Meiyan}~\snm{Li}},
\author[D]{\fnms{Xingtao}~\snm{Zhou}},
\author[B]{\fnms{Bo}~\snm{Fu$^{\ddag}$}\ead[label=e1]{fu@fudan.edu.cn}}, 
\author[A]{\fnms{Catherine C.}~\snm{Liu$^{\ddag}$}\ead[label=e2]{macliu@polyu.edu.hk}
}, 
\and
\author[C]{\fnms{A.H.}~\snm{Welsh}}

\address[A]{Department of Applied Mathematics, The Hong Kong Polytechnic University
\printead[presep={,\ }]{e2}}

\address[B]{School of Data Science, 
Fudan University \printead[presep={,\ }]{e1}}

\address[C]{College of Business and Economics, Australian National University
}
\address[D]{Eye Institute and Department of Ophthalmology, Eye \& ENT Hospital, Fudan University
}
 \end{aug}

\begin{abstract}

The ultra-widefield (UWF) fundus image is an attractive 3D biomarker in AI-aided myopia screening because it provides much richer myopia-related information. 
Though axial length (AL) has been acknowledged to be highly related to the two key targets of myopia screening, Spherical Equivalence (SE) measurement and high myopia diagnosis, its prediction based on the UWF fundus image is rarely considered. 
To save the high expense and time costs of measuring SE and AL, we propose the Copula-enhanced Convolutional Neural Network (CeCNN), a one-stop UWF-based ophthalmic AI framework to jointly predict SE, AL, and myopia status. 
The CeCNN formulates a multiresponse regression that relates multiple dependent discrete-continuous responses and the image covariate, where the nonlinearity of the association is modeled by a backbone CNN.
To thoroughly describe the dependence structure among the responses, we model and incorporate the conditional dependence among responses in a CNN through a new copula-likelihood loss. 
We provide statistical interpretations of the conditional dependence among responses, and reveal that such dependence is beyond the dependence explained by the image covariate. 
We heuristically justify  that the proposed loss can enhance the estimation efficiency of the CNN weights.  
We apply the CeCNN to the UWF dataset collected by us and demonstrate that the CeCNN sharply enhances the predictive capability of various backbone CNNs. 
Our study evidences the ophthalmology view that besides SE, AL is also an important measure to myopia.
\end{abstract}

\blfootnote{$\dag$: Co-first authors. }
\blfootnote{$\ddag$: Co-corresponding authors. }
\end{frontmatter}
\section{Introduction}
{
An important trend in ophthalmology research is to apply deep learning (DL) techniques to fundus images to aid the diagnosis and assessment of ophthalmological diseases 
(\cite{cen2021automatic, kim2021development, li2021deep}; among others). 
There are two main types of fundus images: traditional  fundus images and advanced \textit{ultra-widefield (UWF) fundus images} \citep{midena2022ultra}.
Compared to the former that measures a narrow visual range of 30$^{\circ}$--75$^{\circ}$ (the orange dashed circle in Figure \ref{fig:UWF_vs_regular}), the UWF fundus images offer a much broader 200$^{\circ}$ view of the fundus.  These images are much more informative in \textit{myopia screening} (e.g. the yellow circles in Figure \ref{fig:UWF_vs_regular} reflect lesions associated with myopia that are outside the traditional image), although they require more advanced equipment. 
\begin{figure}[!htb]
    \centering
    \includegraphics[scale = .075]{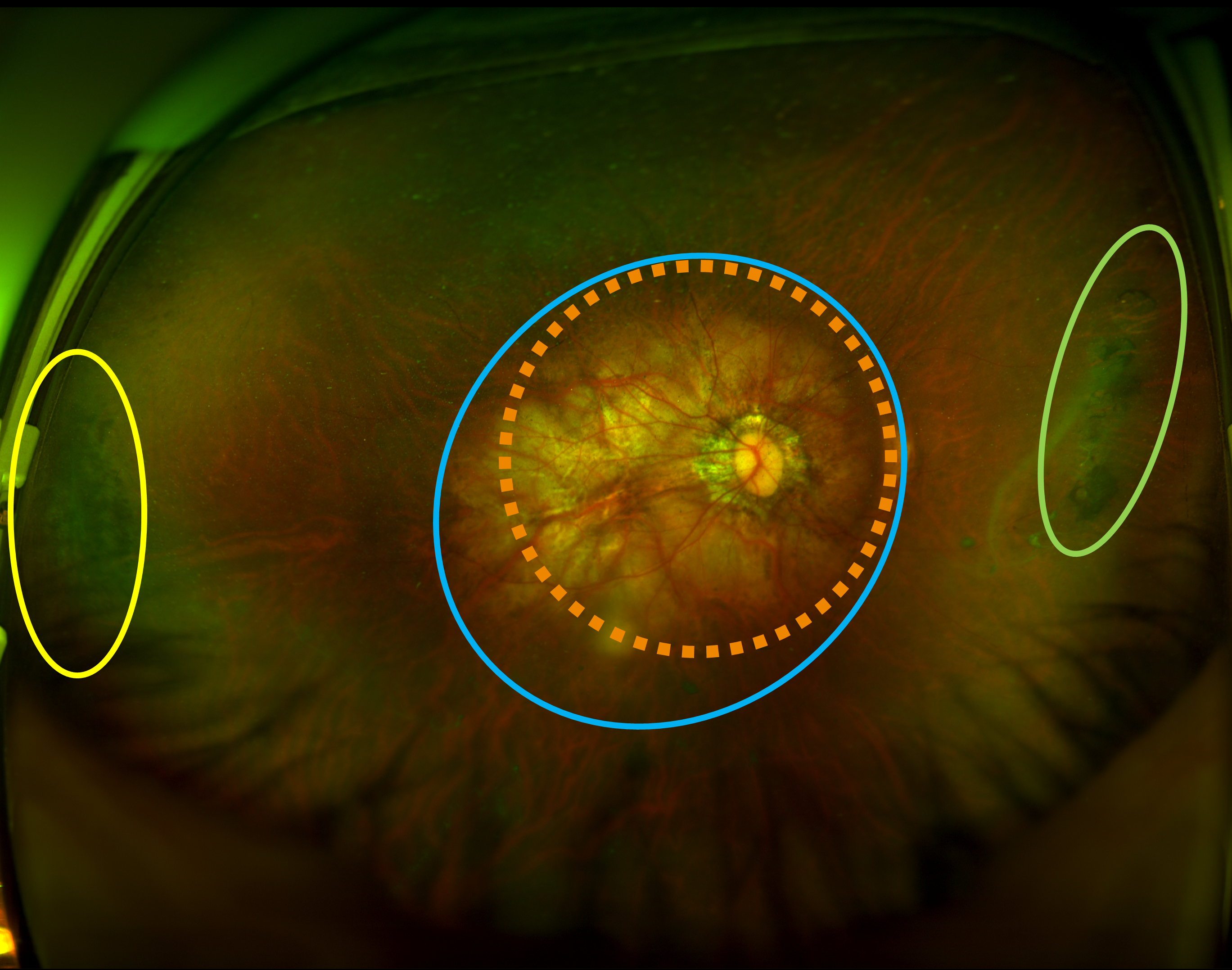}
    \caption{Advantages of UWF imaging in myopia-related pathology. The orange dashed circle represents the area covered by regular fundus images. The area within the yellow circle indicates lesions caused by peripheral laser spots, the region within the blue circle shows extreme peripheral chorioretinal atrophy, and the area within the green circle contains pigmentary degeneration lesions. 
    }
    \label{fig:UWF_vs_regular}
\end{figure}

In myopia screening, the spherical equivalence (SE) acts as the gold standard for the degree of myopia; the larger the magnitude of SE, the higher the myopia status; the cut-off of for high myopia is -8.0 dioptres \citep{kobayashi2005fundus}. 
\textit{High myopia status} is another important concern in myopia screening because high myopia can substantially increase the risk of blindness \citep{iwase2006prevalence}.
Ophthalmological practitioners have also recognised that the axial length (AL) may be meaningful to myopia screening, since AL is a crucial ocular component which combines information on anterior chamber depth, lens thickness, and vitreous chamber depth (\cite{meng2011axial, tideman2016association}). 
Therefore, we are motivated to develop a one-stop scheme that jointly predicts SE and AL and diagnoses high myopia status based on UWF fundus images. 
}

{
\subsection{Motivations} 
There are two motivations for the present study: the ophthalmic need to integrate AL information into myopia prediction and our desire to model conditional dependence among responses in convolutional neural networks (CNN). 

\subsubsection*{Motivation 1: integrating AL into myopia prediction}
The existing ophthalmology literature usually employs AL to predict SE or high myopia status (\cite{mutti2007refractive, haarman2020complications, zhang2024axial}), indicating that integrating  information from AL should enhance myopia prediction. 
Nonetheless, precisely measuring AL in practice is costly and time-consuming \citep{oh2023deep}. 
This drives us to jointly predict SE and AL from the UWF fundus image biomarker. 
Specifically, we relate the bivariate responses to a tensor object (image) covariate through a CNN, the most widely used DL technique for multi-task learning in computer vision. 

\subsubsection*{Motivation 2: modeling conditional dependence among responses in a CNN}
A CNN naturally incorporates the dependence among responses that is explained by the common features from the image covariate. 
However, it is unclear whether a CNN learns the remaining unexplained \textit{conditional dependence among responses given the image covariate}. 
Generally, a CNN is trained under an empirical loss that is the sum of mean squared error (MSE) losses or the sum of cross entropy losses for regression and classification tasks respectively. 
Such empirical losses treat the responses as conditionally independent given the image covariate; refer to our discussion in Section \ref{sec: rethinking} for more details. 
Such a conditional independence assumption may be violated in practice.
In our application, there is a \textit{strong correlation} between SE and AL (see Figure \ref{fig:scatter_plot}) and 
\begin{figure}[!htb]
    \centering
    \includegraphics[scale = 0.35]{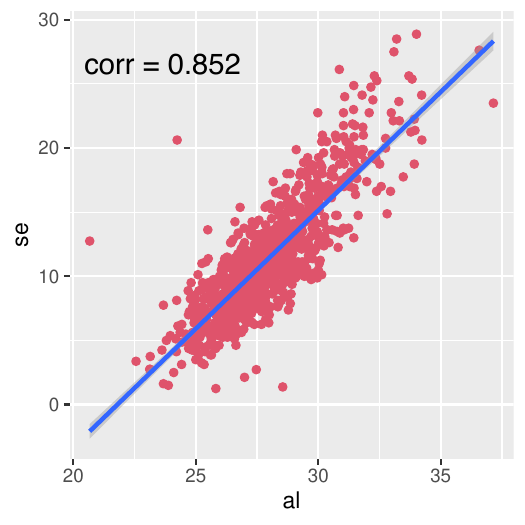}
    \includegraphics[scale = 0.35]{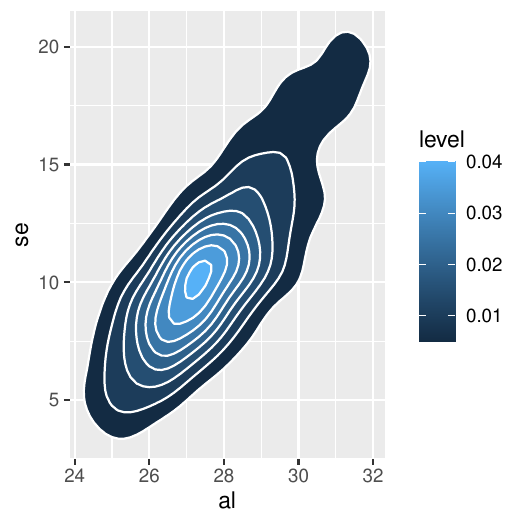}
    \caption{Left: scatter plot; right:contour plot; x axis: AL; y axis: SE. }
    \label{fig:scatter_plot}
\end{figure}
this strong correlation may not be completely explained by the UWF fundus image covariate. 
This drives us to model, interpret, and incorporate the conditional dependence among responses given the image covariate into a CNN within the context of multi-task learning for the purpose of enhancing the prediction of myopia. 
}

\subsection{Related work}
 
{
\subsubsection*{Multiresponse learning in statistics}
In multiresponse models, multivariate classification and regression tree (CART) and its variants are widely used (\cite{de2002multivariate}; \cite{loh2013regression}; \cite{rahman2017integratedmrf}; among others), 
although they do not take the dependence among responses into consideration. 
To model the dependence among responses, copula models \citep{sklar1959fonctions} have been connected with various joint regression analyses (\cite{song2009joint}; \cite{panagiotelis2012pair}; \cite{yang2019nonparametric}; among others). 
However, the methodologies in the above literature only study the influence of scalar covariates on multiple responses and may not be applicable to image (tensor) covariates. 
Recently, there has been some literature studying multiresponse tensor regression models (\cite{raskutti2019convex}; \cite{chen2019non}; \cite{zou2022estimation}; among others);
such works tend to ignore one or more of the nonlinear association, the dependence among responses, and the mixture of discrete-continuous responses.

\subsubsection*{Studies on dependence among responses in deep learning}
In the field of computer vision, the existing DL literature rarely considers the conditional dependence among multiple responses given the image covariate. 
In multi-task learning, most of the literature either uses a simple empirical loss, or uses a weighted sum of marginal MSE or cross entropy losses (\cite{kendall2018multi}; \cite{lin2019pareto}; among others). 
{\it 
Neither of these two practices incorporates the conditional dependence structure of the responses.}
Meanwhile, those weights assigned to each marginal loss may not be well-explained and may be difficult to learn from the data. 
In multi-instance learning, the existing literature considers the spatial correlation between labels determined by small patches/instances of the image covariates (\cite{song2018deep, lai2023single}; among others). 
In multi-label learning, the existing literature models the partial correlation among labels to guide the information propagation through graph convolutional networks (\cite{chen2019multi, sun2022multi}; among others). 
}

\subsection{CeCNN: Copula Enhanced CNN framework}
{
Let $\bm{Y} = (y_{1}, \ldots, y_{ p_1}, y_{ p_1+1}, \ldots, y_{p_1 +p_2})$ be a $(p_1+p_2)$-dimensional response vector that allows both continuous and binary entries.
We call such responses mixed-type responses. 
Without loss of generality, assume $y_{j} \in \mathbb{R}$ for $j \le p_1$ and $y_{j} \in \{0, 1\}$ for $p_1 < j \le p_1+p_2$. 
That is, the pending ophthalmological multi-task learning problem includes $p_1$ regression tasks and $p_2$ binary classification tasks simultaneously. 
Let $\mathcal{X} \in \mathbb{R}^{k_1\times \cdots \times k_J}$ be a $J$th-order tensor, and $\mathcal{G}: \mathbb{R}^{k_1\times \cdots \times k_J} \to \mathbb{R}^{p_1} \times \{0, 1\}^{p_2}$ be an unknown \textit{nonlinear} function that maps a high order tensor to a $(p_1+p_2)$-dimensional outcome. 
We formulate the following multiresponse mean regression that associates responses $\bm{Y}$ with a tensor object covariate $\mathcal{X}$
\begin{align}\label{mod:basic}
    E(\bm{Y}|\mathcal{X}) := \mathcal{G}  (\mathcal{X})= \{g_1(\mathcal{X}), \ldots, g_{p_1}(\mathcal{X}), \mathcal{S} \circ g_{p_1+1}(\mathcal{X}), \ldots, \mathcal{S} \circ g_{p_1+p_2}(\mathcal{X})\}, 
\end{align} 
where $g_j: \mathbb{R}^{k_1\times \cdots \times k_J} \to \mathbb{R}$ are unknown single output non-linear functions for $j=1, \ldots, (p_1+p_2)$, $\circ$ denotes composition of two functions, and $\mathcal{S}(z) = 1/(1+e^{-z})$ is the sigmoid function that maps the real line to $(0, 1)$. 

We model the unknown nonlinear regression functions $\mathcal{G}$ through a backbone CNN with $(p_1 + p_2)$ outputs. 
Although there are various types of backbone CNNs such as LeNet \citep{lecun1998gradient}, ResNet \citep{he2016deep}, and DenseNet \citep{huang2017densely}, they share a similar architecture. 
Usually, the architecture of a CNN consists of many hidden layers, including convolution, pooling, and fully connected layers \citep{lecun2015deep}.}
Fitting a CNN is then equivalent to optimizing a specific loss over the parameters contained in these hidden layers.  
{
In this paper, we propose a new \textit{copula-likelihood loss} to train backbone CNNs with mixed-type outputs. 
Specifically, to meet the emergent needs of ophthalmology practice, we focus on two sets of ophthalmic AI tasks arising from myopia screening, the regression-classification (R-C) task and the regression-regression (R-R) task, 
and derive the form of the proposed copula-likelihood loss in each task respectively. 
The R-C task aims to jointly predict the AL and diagnose the high myopia. 
The proposed loss and the accompanying training procedure create a new AI framework called Copula-enhanced CNN (CeCNN). 
From a statistical perspective, we attempt to interpret the conditional dependence modeled by CeCNN and justify the enhancement in estimation brought by the proposed loss. 
}

{
\subsection{Our contributions}
In this paper, motivated by incorporating AL into myopia screening to enhance myopia prediction, we present a nonlinear multiresponse regression where SE (or the Bernoulli variable of myopia status) and AL are associated with UMF fundus images (mode-3 tensors) through a backbone CNN.
Specifically, we train the backbone CNN by optimizing a proposed \textit{copula-likelihood loss} so as to accommodate the \textit{conditional dependence} among responses that may not be captured by the backbone CNN itself.
We now highlight our main contributions.

Our contributions are tri-fold. 
In ophthalmology, we might be the first to jointly predict SE and AL in myopia screening based on UWF fundus images. 
The present study allows for one-stop measurement of SE, AL, and diagnosis of high myopia through one scan, saving manpower and time costs for the precise measurement of SE and AL. 
Numerical experiments demonstrate that by incorporating the conditional correlation between SE and AL, our method enhances myopia prediction. 
In this sense, our study might be seen as providing the first evidence of the ophthalmological view that besides SE, AL is also an important measure of myopia.

In deep learning, we contribute a new loss which might be the first to model and use the conditional dependence among responses given the image predictor.  
{
We show that the traditional CNN with empirical loss naturally learns the dependence contributed by the common image predictor, but ignores the conditional dependence among responses. 
In contrast, our proposed loss captures both dependencies; refer to equations \eqref{Covariance: R-C} and \eqref{covariance: R-R} for illustration. }
Numerical results demonstrate that the proposed loss leads to better predictive performance compared to the empirical loss and the uncertainty loss for multi-task learning \citep{kendall2018multi}. 
It is anticipated that the proposed loss can be applied to other similar multi-task applications in computer vision. 

In statistics, we might be the first to apply a CNN to relate multiple mixture-type responses to a tensor object covariate nonlinearly and also model the dependence among responses thoroughly. 
We show heuristically that optimizing the proposed copula-likelihood loss leads to lower estimation risk for the CNN weights in the asymptotic setting. 
Our study illustrates statistics harnessing AI through 
\textit{extracting more information within data objects}. 
}

The rest of the paper is organized as follows. 
Sections \ref{sec:reg-clas} and \ref{sec:reg-reg} introduce how the CeCNN works in regression-classification (R-C) and regression-regression (R-R) tasks, respectively. 
{Section \ref{sec: rethinking} rethinks the copula-likelihood loss from the perspective of higher relative efficiency in estimation of CNN weights.} 
Section \ref{sec:app} presents the performance of the CeCNN in myopia prediction on our UWF fundus image dataset. 
Section \ref{sec:sim} carries out simulations on synthetic datasets for illustration. 
Section \ref{sec:discussion} concludes the paper with brief discussions. 
The code for reproducibility is available on GitHub \hyperlink{https://github.com/Charley-HUANG/CeCNN}{https://github.com/Charley-HUANG/CeCNN}.

\section{Regression-classification task}
\label{sec:reg-clas}
We start with our R-C task in myopia screening. 
That is, under model \eqref{mod:basic}, the response vector is $\bm{Y} = (y_{1}, y_{2})$, where $y_{1} \in \mathbb{R}_+$ and $y_{2} \in \{0, 1\}$ denote  the AL and the status of high myopia (1: high myopia or greater than eight diaoptres; 0, otherwise),  respectively. 
The explanatory variable is $\mathcal{X} \in \mathbb{R}^{224 \times 224 \times 3}$,  a UWF fundus image stored in red and green with  $224 \times 224$ channel-wise pixels. 
We construct the copula-likelihood loss for the R-C task in subsection \ref{subsec:R-C loss} and 
summarize the whole CeCNN procedure for the R-C task in subsection \ref{subsec:CeCNN}. 
To characterize the joint distribution of the mixed-type responses $(y_1, y_2)$, we adopt the commonly used copula model \citep{sklar1959fonctions}, which models a joint distribution through a copula and the marginal distributions. 

{
Without loss of generality, a $2$-dimensional ($p$-dim) copula $\mathbb{C}$ is a distribution function on $[0, 1]^2$, where each univariate marginal distribution is uniform on $[0, 1]$. 
One can always express a joint distribution $F$ through a copula $\mathbb{C}$ and the marginal distributions as
$$
F(y_1, y_2) = \mathbb{C}\{F_1(y_1), F_2(y_2)\},
$$
where $F_j$ denotes the $j$th marginal CDF of $y_j$ for $j=1, 2$. 
Let $\Phi$ be the CDF of the standard normal distribution $N(0, 1)$.
The joint CDF under a \textit{Gaussian copula} is
\begin{align*}
  F(y_1, y_2) = \mathbb{C}(\bm{y}|\Gamma) = \Phi_2(\Phi^{-1}\{F_1(y_1)\}, \Phi^{-1}\{F_2(y_2)\}|\Gamma), ~ ~\Gamma = \left(
  \begin{matrix}
      1 & \rho \\
      \rho & 1
  \end{matrix}
  \right),  
\end{align*}
where $\Phi_2(\cdot|\Gamma)$ denotes the CDF of the $2$-dimensional ($p$-dim) Gaussian distribution 
$\text{MVN}(\bm{0}_2, \Gamma)$ with correlation matrix $\Gamma \in \mathbb{R}^{2 \times 2}$, and $\rho \in (-1, 1)$ characterizes the dependence between $(y_1, y_2)$. 
In the presence of covariate $\mathcal{X}$, then the conditional joint CDF given $\mathcal{X}$ is naturally written as
\begin{align}
\label{Gaussian Copula}
    F(y_1, y_2|\mathcal{X}) = \mathbb{C}(\bm{y}|\Gamma, \mathcal{X}) = \Phi_2(\Phi^{-1}\{F_1(y_1|\mathcal{X})\}, \Phi^{-1}\{F_2(y_2|\mathcal{X})\}|\Gamma). 
\end{align}
When both $y_1$ and $y_2$ are continuous, the closed form of the joint density function $f(y_1, y_2|\mathcal{X})$ is straightforward; refer to expressions  \eqref{non-Gaussian joint} and \eqref{likelihood}  in Section \ref{sec:reg-reg}. 
In this section, we derive the closed form of the joint density when $y_1 \in \mathbb{R}$ and $y_2 \in \{0, 1\}$. 
}

\subsection{Copula-likelihood loss}
\label{subsec:R-C loss}
{We begin by modeling the marginal distributions $F_1(y_{1}|\mathcal{X})$ and $F_2(y_2|\mathcal{X})$. 
Note that the contour plot Figure \ref{fig:scatter_plot} (b) yields that $y_1$ is approximately normal on its margin. 
Meanwhile, $y_2$ is naturally Bernoulli.
Therefore, under a CNN $\mathcal{G} = \{g_1, \mathcal{S}\circ g_2\}, $we model the marginal distributions of $(y_1, y_2)$ given $\mathcal{X}$ as 
\begin{align}
    \label{Marginal R-C}
  y_{1}|\mathcal{X} \sim N(g_1(\mathcal{X}), \sigma^2), \quad   y_{2}|\mathcal{X} \sim \text{Bernoulli}[\mathcal{S} \circ g_2(\mathcal{X})]. 
\end{align}
}
{
Let $n$ be the size of the training data. 
For $i=1, \ldots,n$, let $\mu_{i1} = g_1(\mathcal{X}_i)$ be the marginal expectation of $y_{i1}$ given $\mathcal{X}_i$, $\mu_{i2} \equiv Pr\{y_{i2} = 1 |\mathcal{X}_i\} = \mathcal{S} \circ g_2(\mathcal{X}_i)$ be the marginal probability of $y_{i2} = 1$ given $\mathcal{X}_i$, and $z_{i1} = (y_{i1} - \mu_{i1})/\sigma$ be the standardized residual of $y_{i1}$. 
Let $u_{i1} = F_2(y_{i2} - )$ be the left-hand limit of the CDF $F_2$ at $y_{i2}$ and $u_{i2} = F_2(y_{i2})$. 
Let $N(\cdot|\mu, \sigma^2)$ denote the density of $N(\mu, \sigma^2)$. 
The joint density for bivariate discrete-continuous variables $(y_1, y_2)$ given $\mathcal{X}_i$ is
\begin{align}
\label{R-C density}
    f(y_{i1}, y_{i2}|\mathcal{X}_i) = N(y_{i1}|\mu_{i1}, \sigma^2) \sum_{r=1}^2 (-1)^r \mathbb{C}_1^2 (z_{i1}, u_{ir}|\Gamma), 
\end{align}
where $\mathbb{C}_1^2$ is the partial derivative of the Gaussian copula \eqref{Gaussian Copula} with respect to the continuous coordinate $y_{i1}$. 

Specifically, for $y_{i2} = 1$ in our case, up to a normalized constant $c_0$, we have 
\begin{eqnarray*}
    \begin{aligned}
   \mathbb{C}_1^2(z_{i1}, u_{i1}|\Gamma) &= c_0 \frac{1}{\sqrt{1-\rho^2}} \int_{-\infty}^{\Phi^{-1}(1-\mu_{i2})} \exp\left\{-\frac{1}{2} (z_{i1}, s)\Gamma^{-1} (z_{i1}, s)^{T} + \frac{1}{2} z_{i1}^2\right\}ds \\
& = c_0\frac{1}{\sqrt{1-\rho^2}} \int_{-\infty}^{-\Phi^{-1}(\mu_{i2})} \exp\left\{-\frac{z_{i1}^2 - 2\rho z_{i1}s + s^2}{2(1-\rho^2)} + \frac{z_{i1}^2}{2}\right\}ds \\
& = \Phi\left( \frac{-\Phi^{-1}(\mu_{i2}) - \rho z_{i1}}{\sqrt{1-\rho^2}}\right) = 1- \Phi\left(\frac{\Phi^{-1}(\mu_{i2}) + \rho z_{i1}}{\sqrt{1-\rho^2}}\right),  \\
    \end{aligned}
\end{eqnarray*}
and 
\begin{align*}
   \mathbb{C}_1^2(z_{i1}, u_{i2}|\Gamma) &= c_0\frac{1}{\sqrt{1-\rho^2}} \int_{-\infty}^{+\infty} \exp\left\{-\frac{1}{2} (z_{i1}, s)\Gamma^{-1} (z_{i1}, s)^{T} + \frac{1}{2} z_{i1}^2\right\}ds  = 1. 
\end{align*}
Consequently, we have the conditional probability
\begin{align*}
    Pr\{y_{i2} = 1 |  y_{i1} = \sigma z_{i1} + \mu_{i1}\} = \Phi\left(\frac{\Phi^{-1}(\mu_{i2}) + \rho z_{i1}}{\sqrt{1-\rho^2}}\right) \equiv C^*(\mu_{i2}, z_{i1}|\rho). 
\end{align*}
Hence, by taking the value of $y_{i2}$ to be either $0$ or $1$, corresponding to \eqref{R-C density}, the closed form of the conditional joint density of $(y_{i1}, y_{i2})$ given $\mathcal{X}_i$ in our case is 
\begin{align*}
    l(y_{i1}, y_{i2}|\mathcal{X}_i) 
    = N(y_{i1}; g_1(\mathcal{X}_i), \sigma^2) C^*(\mu_{i2}, z_{i1}|\rho)^{y_{i2}} (1- C^*(\mu_{i2}, z_{i1}|\rho))^{1-y_{i2}}. 
\end{align*}
Finally, we obtain the copula-likelihood loss, which is  minus the log-likelihood for the training data
\begin{eqnarray}
\label{R-C loss}
    \begin{aligned}
        \mathcal{L}_1 &(g_1, g_2|\{\bm{Y}_i\}_{i=1}^n, \{\mathcal{X}_i\}_{i=1}^n,  \rho,  \sigma) = \frac{1}{2\sigma^2} \sum_{i=1}^n (y_{i1} - \mu_{i1})^2\\ 
        &- \left\{\sum_{i=1}^n [y_{i2} \log C^*(\mu_{i2}, z_{i1}|\rho)
        + (1-y_{i2}) \log\{1- C^*(\mu_{i2}, z_{i1}|\rho)\}] \right\}. 
    \end{aligned}
\end{eqnarray}
Note that on the right hand side (RHS) of \eqref{R-C loss}, the first summand is related to $g_1$ only, while the second summand  associates $\mu_{i1}$ and $\mu_{i2}$, or equivalently, $g_1$ and $g_2$ through a parameter $\rho$. 
Mathematically, when $\rho = 0$, $C^*(\mu_{i2}, z_{i1}|\rho)$ becomes $\mu_{i2}=\mathcal{S} \circ g_2(\mathcal{X}_i)$, implying that the second summand in the RHS of equation \eqref{R-C loss} reduces to a pure cross entropy loss of $g_2$ only. 
Thus, loss \eqref{R-C loss} is a more general form of the empirical loss. 
{
Both $\rho$ and $\sigma$ have explicit statistical interpretations. 
The scale parameter $\sigma$ represents the standard deviation of $y_1|\mathcal{X}$, 
acting as the weight balancing the MSE loss and the cross-entropy-like loss. 
The interpretation of $\rho$ in the R-C task is given by the following theorem. 
}
}
{
\begin{theorem}
\label{theo: distribution of Gaussian scores}
Suppose the joint distribution of  $\bm{Y} = (y_1, y_2)$ is given by Gaussian copula \eqref{Gaussian Copula} with correlation matrix $\Gamma$, where the marginal distributions of $y_1$ and $y_2$ are given by \eqref{Marginal R-C}. 
Let $\mu_1 = g_1(\mathcal{X})$ and $\mu_2 = \mathcal{S} \circ g_2(\mathcal{X})$ be the conditional mean of $y_1$ and $y_2$ given image covariate $\mathcal{X}$, respectively.  
Let $z_1|\mathcal{X} = (y_1 - \mu_1)/\sigma$ be the standardized version of $y_1|\mathcal{X}$.
Let $z_2|\mathcal{X} \sim N(\Phi^{-1}(\mu_2), 1)$ be a latent Gaussian score such that $
    (z_{1}, z_{2})^T|\mathcal{X} \sim \text{MVN}\{[0, \Phi^{-1}(\mu_2)]^T, \Gamma\}. $
Then equivalently we have $y_2 = I(z_2 > 0)$. 
\end{theorem}

Theorem \ref{theo: distribution of Gaussian scores} expresses the discrete response $y_2$ in terms of the continuous latent Gaussian score $z_2$ under the Gaussian copula \eqref{Gaussian Copula}. 
It is validated by showing that $Pr\{y_2=1|\mathcal{X}\} = Pr\{z_2>0|\mathcal{X}\}$ and $Pr\{y_2=1|\mathcal{X}, z_1\} = Pr\{z_2>0|\mathcal{X}, z_1\}$ in Supplement A.1. 
As a result, the theorem indicates that, the association matrix $\Gamma$ of the Gaussian copula \eqref{Gaussian Copula} becomes \textit{the correlation matrix of the conditional joint distribution of $z_1$ (standardized $y_1$) and the latent Gaussian score $z_2$ given the image $\mathcal{X}$.}
Our result is motivated by the Gaussian score correlation \citep[Definition 6.3]{peter2007correlated} that characterizes the association between two continuous variables. 
In the presence of a Bernoulli variable $y_2$, we construct a continuous latent variable $z_2$ as a replacement for $y_2$ that maintains the Gaussian score correlation interpretation.

From Theorem \ref{theo: distribution of Gaussian scores}, after some algebra, the full covariance structure between the two Gaussian variables $(y_1, z_2)$ is 
\begin{align}
    \label{Covariance: R-C}
    \text{Cov}(y_1, z_2) = \rho \sigma +  \text{Cov}(g_1(\mathcal{X}), \Phi^{-1}\{\mathcal{S} \circ g_2(\mathcal{X})\}). 
\end{align}
The latter summand in the RHS, $\text{Cov}(g_1(\mathcal{X}), \Phi^{-1}\{\mathcal{S} \circ g_2(\mathcal{X})\})$, is  determined by the image $\mathcal{X}$ and is caught by the CNN. 
However, the conditional correlation $\rho$ between $y_1$ and $z_2$ given $\mathcal{X}$, is beyond the dependence/correlation explained by the image covariate. 
}

{
The covariance structure \eqref{Covariance: R-C} inspires natural estimators of $\rho$ and $\sigma$. 
Once the marginal estimators of $g_1$ and $g_2$ denoted as $\hat{g}_1^0$ and $\hat{g}_2^0$ respectively, are obtained, one may, i) estimate $\rho$ as the Pearson correlation between $y_1$ and $\Phi^{-1}\{\mathcal{S} \circ \hat{g}_2^0(\mathcal{X})\}$, by treating $\Phi^{-1}\{\mathcal{S} \circ \hat{g}_2^0(\mathcal{X})\}$ as a realization of $z_2|\mathcal{X}$; 
ii) estimate the scale parameter $\sigma$ as the standard deviation of the residuals of $\hat{g}_1^0(\mathcal{X})$. 
In summary, let $\bm{\mathcal{X}}$ denote all the  images $\mathcal{X}$ in the training set. 
The estimators of the copula parameters $(\rho, \sigma)$ are
\begin{align}
    \label{R-C estimation}
    \hat{\rho} = corr(\bm{y}_1, \Phi^{-1}\{\mathcal{S}\circ \hat{g}_2^0(\bm{\mathcal{X}})\}), ~~\hat{\sigma} = \text{sd}\{\bm{y}_1 - \hat{g}_1^0(\bm{\mathcal{X}})\}. 
\end{align}
Consequently, the proposed copula-likelihood loss \eqref{R-C loss} is specified by using the estimators $(\hat{\rho}, \hat{\sigma})$ in place of the unknown $(\rho, \sigma)$. 

}

\subsection{End-to-end CeCNN}
\label{subsec:CeCNN}
{
In subsection \ref{subsec:R-C loss}, we formulated the proposed copula-likelihood loss accompanied by the estimators of the copula parameters. 
In this subsection, we illustrate how statistics harnesses AI through the proposed CeCNN framework. 
The overall CeCNN framework has three modules, the warm-up CNN, copula estimation, and the C-CNN. 
The warm-up CNN module is basically a backbone CNN trained under the empirical loss, providing the marginal estimators needed for copula parameter estimation. 
The copula estimation module estimates the parameters $(\rho, \sigma)$ based on the marginal estimators obtained by the warm-up CNN. 
The last C-CNN module is the core of the whole CeCNN framework where the backbone CNN is trained under the proposed copula-likelihood loss to incorporate the information of conditional dependence. 
The three modules are summarized in Algorithm \ref{alg: CeCNN RC}.

\begin{algorithm}[!htb]
\scriptsize
\caption{End-to-end CeCNN (regression-classification task)}\label{alg: CeCNN RC}
\begin{algorithmic}[1]
\Require Training images $\bm{\mathcal{X}} = \{\mathcal{X}_i\}_{i=1}^n$ and training labels $\{\bm{Y}_i = (y_{i1}, y_{i2})\}_{i=1}^n$. 
\Ensure CeCNN estimator $\hat{\mathcal{G} }= (\hat{g}_{1}, \mathcal{S} \circ \hat{g}_{2})$. 
\Algphase{Module 1: The warm-up CNN}
\State Design a bivariate-output backbone CNN $\mathcal{G} = (g_1, \mathcal{S} \circ g_2)$, where $g_j = \text{F-C}_j \circ \text{Pool}(1, \ldots, k_2) \circ \text{Conv}(1, \ldots, k_1)$, $\text{F-C}_j(\bm{z}) = \bm{w}_{j}^T \bm{z} + b_j$, for $j=1, 2$. 
\State Obtain marginal estimator $\hat{g}_{1}^0 = \arg \min_{g_1}n^{-1} \sum_{i=1}^n (y_{i1} - g_1(\mathcal{X}_i))^2$. 
\State Obatin marginal estimator $\hat{g}_{2}^{0} = \arg \min_{g_2} -\sum_{i=1}^n [y_{i2}\log(\mathcal{S} \circ g_2(\mathcal{X}_i)) + (1-y_i) \log(1- \mathcal{S} \circ g_2(\mathcal{X}_i))]$. 
\Algphase{Module 2: Copula estimation}
\State Obtain estimators $(\hat{\rho}, \hat{\sigma})$ of the copula parameters by \eqref{R-C estimation}. 
    \Algphase{Module 3: The C-CNN}
\State Determine the copula-likelihood loss $\mathcal{L}_1(g_1, g_2| \mathcal{X}, \bm{Y}, \hat{\rho}, \hat{\sigma})$ in the form of \eqref{R-C loss}. 
\State Obtain $\hat{\mathcal{G}} = (\hat{g}_1, \hat{g}_2) = \arg \min \limits_{g_1, g_2} \mathcal{L}_1$. 

\end{algorithmic}
\end{algorithm}

In Module 1, without loss of generality, we assume that the backbone CNN $\mathcal{G}$ has $k_1$ convolution (Conv) layers, $k_2$ pooling (Pool) layers, and one fully connected (F-C) layer (e.g. the LeNet and the ResNet backbone CNNs). 
The regression and classification tasks differ only in the F-C layer and share the Conv and Pool layers. 
All the numerous parameters included in the very deep hidden layers are updated by the Adam algorithm \citep{kingma2014adam} to optimize the empirical losses presented in lines 2 and 3, respectively, until convergence.

We view the outputs of the warm-up CNN as the marginal estimators $\hat{g}_1^0$ and $\hat{g}_2^0$ ($\hat{g}_2^0$ is the classification output before the sigmoid transformation). 
Then in Module 2, we obtain estimators $(\hat{\rho}, \hat{\sigma})$ following \eqref{R-C estimation} and thus determine the copula-likelihood loss \eqref{R-C loss}. 
In Module 3, we fix $(\hat{\rho}, \hat{\sigma})$ and only update the weights in the backbone CNN. 
This module looks as if a fine-tuning on the pre-trained warm-up CNN. 
More detail about the training procedure in Module 3 is given in Section \ref{sec:discussion}. 
}

\section{Regression-regression task}
\label{sec:reg-reg}
This section treats the specific regression-regression (R-R) task of predicting the clinically important, highly correlated responses SE and AL, using the proposed CeCNN. 

Let $\bm{Y} \in \mathbb{R}^p$. 
{
We rewrite model \eqref{mod:basic}  as the following equivalent multiresponse regression  model
\begin{align}
    \label{pureres}
    \bm{Y} = \mathcal{G}(\mathcal{X}) + \bm{\epsilon}, 
\end{align}
where $\bm{\epsilon} = (\epsilon_{1}, \ldots, \epsilon_{p})$ is a $p$-dimensional noise vector,  $\bm{\epsilon} \perp \mathcal{X}$, $E(\epsilon_{j}) = 0$ for all $1\le j \le p$. 
For any $p \ge 2$, \eqref{pureres} expresses multiresponse regression in a unified form. 
{
Expression \eqref{pureres} yields the following covariance structure among $(y_1, \ldots, y_p)$, 
\begin{align}
\label{covariance: R-R}
    \text{Cov}(y_s, y_t) = \text{Cov}(\epsilon_s, \epsilon_t) + \text{Cov}\{g_s(\mathcal{X}), g_t(\mathcal{X})\}, ~~s, t =1, \ldots, p. 
\end{align}
}
{
Here $\text{Cov}(\epsilon_s, \epsilon_t)$ characterizes the \textit{conditional dependence among $\bm{Y}$ given $\mathcal{X}$}, which originates from the model error $\bm{\epsilon}$ and is beyond the images; 
$\text{Cov}\{g_s(\mathcal{X}), g_t(\mathcal{X})\}$ characterizes the dependence of $\bm{Y}$, which is learned by the CNN and contributed by the image $\mathcal{X}$. 

Under \eqref{pureres}, the distribution of $ \bm{Y} - \mathcal{G}(\mathcal{X})|\mathcal{X}$ is the same as that of $\bm{\epsilon}$ and it is notationally convenient to express things in terms of $\bm{\epsilon}$.  
Thus, in Gaussian copula modeling, we first have to specify the marginal CDF and the density of $\epsilon_j$, for $j=1, \ldots, p$, is simpler than stating we have to specify the conditional CDF and density of $Y_j- g_j(\mathcal{X})|\mathcal{X}$. 
In this section, we study two types of marginal densities. 
}
} \\

\noindent{\textbf{Nonparametric error}}
We start from a general case where the marginal density of the error $\epsilon_j$ is unspecified. 
In this case, we estimate the marginal CDF $F_{\epsilon_j}$ (and density $f_{\epsilon_{j}}$) first for $j=1, \ldots, p$, then estimate $\Gamma$, and finally we combine them to derive the copula-likelihood loss. 

We begin with the warm-up CNN equipped with the empirical MSE loss and obtain the residuals $(e_{i1}, \ldots, e_{ip})$ on the training dataset, for $i =1, \ldots, n$. 
To make sure $E(\epsilon_j) = 0$, we use the mean-centered residuals $\Tilde{e}_{i1}, \ldots, \Tilde{e}_{ip}$ to obtain the marginal empirical CDF (and pdf) of $\epsilon_j$ with a Gaussian kernel smoothing 
\begin{align}
    \label{smooth}
    \Tilde{F}_{\epsilon_j}(t) = \frac{1}{n}\sum_{i=1}^n \Phi\{(t - \Tilde{e}_{ij})/\psi_0\} , ~ \Tilde{f}_{\epsilon_j}(t) = \frac{1}{n\psi_0}\sum_{i=1}^n \phi\{(t - \Tilde{e}_{ij})/\psi_0\},
\end{align}
where $\psi_0>0$ is a tuning parameter acting as the bandwidth in density estimation. 

Once we obtain the estimates  $\Tilde{F}_{\epsilon_j}$, the correlation matrix $\Gamma$ in the Gaussian copula \eqref{Gaussian Copula} is estimated using
\begin{align}
\label{R-R nonparametric estimation}
    \hat{\gamma}_{sj} = corr(\{\Phi^{-1}(\Tilde{F}_{\epsilon_s} (\Tilde{e}_{is}))\}_{i=1}^n, \{\Phi^{-1}(\Tilde{F}_{\epsilon_j} (\Tilde{e}_{ij}))\}_{i=1}^n), ~s, j =1, \ldots, p,
\end{align}
the Pearson correlation between the two Gaussian scores of the smoothed empirical CDFs of the centered residuals. 

With the above estimates, under the Gaussian copula, based on \citet[eq. (7)]{song2009joint}, the copula-likelihood loss is given by
{\small
\begin{align}
    \label{non-Gaussian joint}
    \mathcal{L}_2(\{g_j\}_{j=1}^p |\{\bm{Y}_i\}_{i=1}^n; \bm{\mathcal{X}}) = - \sum_{i=1}^n \frac{1}{2} \bm{q}_i^T(\bm{I}_p - \Gamma^{-1})\bm{q}_i -\sum_{i=1}^n \sum_{j=1}^p \log\{\Tilde{f}_{\epsilon_j} (y_{ij} - g_j(\mathcal{X}_i))\}, 
\end{align}
}
where $\bm{q}_i = (\Phi^{-1}\{\Tilde{F}_{\epsilon_1}[y_{i1} - g_1(\mathcal{X}_i)]\}, \ldots, \Phi^{-1}\{\Tilde{F}_{\epsilon_p}[y_{ip} - g_p(\mathcal{X}_i)]\})^T$. 
Obviously, if $\Gamma = \bm{I}_p$ (no correlation between the responses), loss \eqref{non-Gaussian joint} reduces to the sum of minus the log densities $\Tilde{f}_{\epsilon_j}$. \\

\noindent{\textbf{Gaussian error}}~
Since both SE and AL look normally distributed (based on the contour plot in Figure \ref{fig:scatter_plot}), 
we may simply adopt the Gaussian model error $\epsilon_j \sim N(0, \sigma_j^2)$, where $\sigma_j$ is a scale parameter. 
With Gaussian error, the likelihood contribution of $\bm{Y}_i |\mathcal{X}_i$ reduces to the multivariate Gaussian density directly. 
Thus, for training data $(\{\bm{Y}_i\}_{i=1}^n, \bm{\mathcal{X}})$, the copula-likelihood loss is 
{\small
\begin{align}
\label{likelihood}
 \mathcal{L}_3(\{g_j\}_{j=1}^p |\{\bm{Y}_i\}_{i=1}^n; \bm{\mathcal{X}}) = -\sum_{i=1}^n \log \text{MVN}_p\{(y_{i1}- g_1[(\mathcal{X}_i)], \ldots, y_{ip}- g_p[(\mathcal{X}_i)]); \bm{0}_p, \Sigma\},
\end{align}
}
where $\Sigma = diag(\sigma_1, \ldots, \sigma_p) \Gamma diag(\sigma_1, \ldots, \sigma_p) \equiv (\sigma_{tj})_{p \times p}$ is the covariance matrix of $\bm{\epsilon}$. 
 Specifically, in our bivariate application, if $\rho = 0$ (that is, $\Gamma$ is identity), the copula-likelihood loss \eqref{likelihood} simply reduces to the sum of the empirical MSE loss and some constant. 
Therefore, the copula-likelihood loss \eqref{likelihood} is a generalization of the empirical MSE loss. 

With Gaussian errors, we only need to estimate the covariance matrix $\Sigma$, or equivalently, the correlation $\Gamma$ and the marginal standard deviations $(\sigma_1, \ldots, \sigma_p)$.
Intuitively, their estimates can be obtained from the empirical correlation matrix and marginal standard deviations of the residuals from warm-up CNNs. 
Let $(\hat{g}_{01}, \ldots, \hat{g}_{0p})$ be $p$-dimensional outputs of CNNs trained with the empirical MSE loss for $(g_1, \ldots, g_p)$ on the training dataset.  
Using the residuals $(e_{i1}, \ldots, e_{ip})$ for each observation, where $e_{ij} = y_{ij} - \hat{g}_{0j}(\bm{X}_i)$, 
we obtain estimates of the copula parameters $\Gamma$ and $\sigma_j$ as 
\begin{align}
    \label{R-R parametric estimation}
    \hat{\Gamma} \equiv (\hat{\gamma}_{tj})_{p\times p}  = corr(\{e_{it}\}_{i=1}^n, \{e_{ij}\}_{i=1}^n), ~ \hat{\sigma}_j = \text{sd}(\{e_{ij}\}_{i=1}^n).
\end{align}

Finally, we summarize the CeCNN for the regression-regression task in Algorithm \ref{alg:CeCNNreg}. 
In this task, the CeCNN again has the three-module structure with different copula parameters. 
For Gaussian error, the copula parameters are the marginal SDs $\sigma_j$ and the Pearson correlations $\gamma_{tj}$; 
for nonparametric error, the copula parameters contain an infinite dimensional parameter $f_{\epsilon_j}$ and the transformed correlation $\gamma_{tj}$.

\begin{algorithm}[!htb]
\scriptsize
\caption{End-to-end CeCNN (regression-regression task)}\label{alg:CeCNNreg}
\begin{algorithmic}[1]
\Require Training images $\bm{\mathcal{X}} = \{\mathcal{X}_i\}_{i=1}^n$ and training labels $\{\bm{Y}_i = (y_{i1}, \ldots, y_{ip})\}_{i=1}^n$. 
\Ensure CeCNN estimator $\hat{\mathcal{G}} = (\hat{g}_{1}, \ldots, \hat{g}_{p})$. 
\Algphase{Module 1: The warm-up CNN}
\State Design a multi-output backbone CNN $\mathcal{G} = (g_1, \ldots,  g_p)$, where $g_j = \text{F-C}_j \circ \text{Pool}(1, \ldots, k_2) \circ \text{Conv}(1, \ldots, k_1)$, $\text{F-C}_j(\bm{z}) = \bm{w}_{j}^T \bm{z} + b_j$, for $j=1, \ldots, p$. 
\State  Obtain marginal estimators $\hat{g}_{j}^0 = \arg \min_{g_j}n^{-1} \sum_{i=1}^n (y_{ij} - g_j(\mathcal{X}_i))^2$, for $j=1, \ldots, p$. 
\Algphase{Module 2: Copula estimation}
\If{Non-Gaussian}
\State Estimate copula parameter $\Gamma$ based on \eqref{R-R nonparametric estimation}. 
\ElsIf{Gaussian}
\State Estimate copula parameters $(\Gamma, \sigma)$ based on \eqref{R-R parametric estimation}. 
\EndIf
\Algphase{Module 3: The C-CNN}
\If{Non-Gaussian}
\State Define the loss  $\mathcal{L}_2$ as \eqref{non-Gaussian joint}. 
Obtain $\hat{\mathcal{G}} = (\hat{g}_1, \ldots, \hat{g}_{p}) =  \arg \min\limits_{g_1, \ldots, g_p} \mathcal{L}_2$. 
\ElsIf{Gaussian}
\State  Define the loss  $\mathcal{L}_3$ as \eqref{likelihood}. Obtain $\hat{\mathcal{G}} = (\hat{g}_1, \ldots, \hat{g}_{p}) =  \arg \min\limits_{g_1, \ldots, g_p} \mathcal{L}_3$. 
\EndIf

\end{algorithmic}
\end{algorithm}

{
\section{Rethinking the copula-likelihood loss: relative efficiency}
\label{sec: rethinking}
An interesting question raised by the reviewers is what kind of ``dependence'' is learned by the copula-likelihood loss
in addition to the ``dependence'' explained by the image covariates, which is learned automatically by the ``baseline''. 
Here the baseline refers to a reasonable backbone CNN equipped with the empirical loss. 
Our understanding is, the proposed copula-likelihood loss \textit{incorporates the conditional dependence among responses given the image covariate}. 
Such conditional dependence cannot be learned by the backbone CNN equipped with the empirical loss. 
As a result, the estimators of the weights in the last fully-connected (F-C) layer of a CNN are asymptotically more efficient under the copula-likelihood loss than those under the empirical loss.
In the following, we confine our discussion to the bivariate-response learning task, considering our application to the UWF dataset.

In both R-C and R-R tasks, the backbone CNN with the empirical loss can only capture the correlation between $y_1$ and $y_2$ explained by the image covariate $\mathcal{X}$, and misses the conditional dependence structure given $\mathcal{X}$. 
This is evidenced by the fact that the CNN fitted under the empirical loss is the nonparametric maximum likelihood estimator on the space of CNNs under the conditional independence model assumption $y_1 \perp y_2 |\mathcal{X}$, 
that is, the special case $\rho = 0$ in our Gaussian copula modeling; 
referred to Propositions A.1 for R-R tasks and A.2 for R-C tasks in the supplement, respectively. 
Therefore, the baseline with the empirical loss may suffer from the risk of model misspecification if the true $\rho \not = 0$, incurring suboptimal prediction. 

In contrast, the copula-likelihood loss delivers the conditional dependence information to the CNN through the correlation parameter $\rho$ in the Gaussian copula. 
{
Specifically, in the R-R task, the loss incorporates the covariance structure of the model error as expressed in \eqref{covariance: R-R};}
in the R-C task, the loss incorporates $\rho = corr(z_1, z_2|\mathcal{X})$, where $z_1$ is the standardized residual and $z_2$ is the latent Gaussian score defined in Theorem \ref{theo: distribution of Gaussian scores}.

{
Suppose the depth of a backbone CNN is $L = k_1+k_2+ 1$. 
Recall that in Algorithms \ref{alg: CeCNN RC} and \ref{alg:CeCNNreg}, the last F-C layer can be expressed as $H = (\text{F-C}_1, \text{F-C}_2)$; 
the stacking of $(1, \ldots, L-1)$ hidden layers can be expressed as  $D = \text{Pool}(1, \ldots, k_2) \circ \text{Conv}(1, \ldots, k_1)$. 
Therefore, $\mathcal{G} = H \circ D$. 
}
By definition, $D: \mathbb{R}^{a \times b \times c} \to \mathbb{R}^K$ is a nonlinear function mapping a tensor $\mathcal{X}$ to $K$ feature maps, where $K$ is the width of the last F-C layer of the CNN. 
The width $K$ varies among different backbone CNNs (e.g. $K=512$ in ResNet and $K=4096$ in DenseNet). 
With $K$ feature maps $D(\mathcal{X})$, the last F-C layer $H: \mathbb{R}^K \to \mathbb{R}^2$ is defined by
\begin{align}
    \label{F-C layer}
    H \circ D(\mathcal{X})= (\hat{y}_1, \hat{y}_2)^T, ~ \hat{y}_j = \mathcal{A}_j\{\bm{w}_j^T D(\mathcal{X}) + b_j\}, ~j=1, 2, 
\end{align}
where $\mathcal{A}_j$ are some specific activation functions, and $\bm{w}_j \in \mathbb{R}^K$ and $b_j \in \mathbb{R}$ are the weights and bias of the $j$th output neuron, respectively. 
In the R-R task, the two activation functions $\mathcal{A}_j$ are both the identity projection; 
in the R-C task, the $\mathcal{A}_j$ are the identity projection and the sigmoid function, respectively.

Fitting a CNN is equivalent to optimizing some loss between $\hat{y}_j$ in \eqref{F-C layer} and the training responses $y_j$.  
In the R-R task, 
we denote by $\hat{\bm{w}}_j^{emp}$ the estimate of $\bm{w}_j$ under the empirical loss and 
denote by $\hat{\bm{w}}_j^{cop}$ the estimate of $\bm{w}_j$ under the copula-likelihood loss. 
We make the following assumption. 

\begin{assumption}[Uncovered feature maps]
\label{ass: uncover}
    Let $\bm{1}_{\bm{w}_j} = \{k: w_{jk} \not = 0\}$ be the index set of non-zero entries in weights $\bm{w}_j$.
   Assume that $\bm{1}_{\bm{w}_1} \setminus \bm{1}_{\bm{w}_2} \not = \emptyset$ and $\bm{1}_{\bm{w}_2} \setminus \bm{1}_{\bm{w}_2} \not = \emptyset$. 
\end{assumption}
Assumption \ref{ass: uncover} assumes that the two outputs $y_1$ and $y_2$ in the R-R tasks share different features of $\mathcal{X}$ extracted from the convolutional layers. 
This assumption naturally holds for those backbone CNNs that assign different F-C layers to different outputs (e.g \cite{liu2019joint}; \cite{lian2022multi}; among others). 
For general backbone CNNs like ResNet and DenseNet, Assumption \ref{ass: uncover} may be examined by significance tests for black box learners \citep{dai2022significance} or by visual explanations for neural networks \citep{selvaraju2017grad}.

Let $w_{jk}$ be the $p$th element of $\bm{w}_j$, for $k=1, \ldots, K$. 
We are in a position to provide the following theorem on estimation of the weights in the last F-C layer of a CNN. 
{
The proof relies on transforming the linear model \eqref{F-C layer} into a seemingly unrelated regression model \citep{zellner1962efficient}; details are deferred to Supplement A.3. 
The theorem is confined to the R-R task; similar results may also hold on the R-C task but the techniques are much more complicated. }
\begin{theorem}
[Relative efficiency]
\label{theo: lower variance}
Let $\bm{\mathcal{X}}$ be the samples of the image covariates in the training set. 
    In the R-R task, under Assumptions \ref{ass: uncover} and other technical assumptions in Supplement A, if  $\bm{1}_{\bm{w}_1} \bigcap \bm{1}_{\bm{w}_2} = \emptyset$, 
    as $n \to \infty$, for $k=1, \ldots, K$, we have 
    $$
    Pr\{\text{Var}(\hat{w}_{jk}^{cop}|\bm{\mathcal{X}}) \le \text{Var}(\hat{w}_{jk}^{emp}|\bm{\mathcal{X}})\} \to 1, ~j=1, 2.  
    $$
\end{theorem}
Note that both $\text{Var}(\hat{w}_{jk}^{cop}|\bm{\mathcal{X}})$ and $\text{Var}(\hat{w}_{jk}^{emp}|\bm{\mathcal{X}})$ are determined by the feature maps of the training image samples and thus they are indeed stochastic. 
Theorem \ref{theo: lower variance} tells that in the asymptotic setting, the former is dominated by the latter. 
Since both $\hat{w}_{jk}^{cop}$ and $\hat{w}_{jk}^{emp}$ are unbiased for $w_{jk}$ (refer to Proposition A.3 in the supplement), Theorem \ref{theo: lower variance} indicates that, with probability tending to 1, $\hat{w}_{jk}^{cop}$ is  more efficient than $\hat{w}_{jk}^{emp}$. 
Therefore, compared with the empirical loss, in the R-R task, the copula-likelihood loss reduces $\text{E}||\bm{w}_j - \hat{\bm{w}}_j||_2^2$, 
the estimation risk of the weights within the $j$th output neuron in the last F-C layer of a CNN.}

\section{Application to the UWF fundus image dataset}
\label{sec:app}
We apply the proposed CeCNN to myopia screening based on our UWF dataset. 
To evaluate the predictive capability of CeCNN, we conduct 10 rounds of 5-fold cross validation on the dataset. 

\subsection{Data preparation}
The data collection process involved capturing 987 fundus images from the left eyes of 987 patients using the Optomap Daytona scanning laser ophthalmoscope (Daytona, Optos, UK).
All enrolled patients sought refractive surgery treatment and were exclusively myopia patients. To ensure homogeneity and accuracy, individuals with other ocular conditions such as cataract, vitreoretinal diseases, or glaucoma, as well as those with a history of trauma or previous ocular surgery, were excluded from the dataset. 
For image selection, we  required the fovea to be positioned at the center of the image and applied the criterion of gradability. Images were considered gradable if there was no blurring in the optic disc or foveal area, and if less than 50\% of the peripheral retinal area was obscured by eyelids or eyelashes. The UWF fundus images obtained during the study were exported in JPEG format and compressed to a resolution of $224 \times 224$ pixels to facilitate subsequent analysis.\\

\noindent{\textbf{Response variables}}
{
As stated before, the response variables considered include two continuous responses, SE and AL, and a binary myopia status response indicating whether the patient has high myopia or not. }
The ground truth values for AL and SE in our study are obtained through standard clinical procedures by professional ophthalmologists from the Eye \& ENT Hospital of Fudan University, ensuring that our ground truth values are accurate and reliable.
High myopia is defined according to SE value with a cut-off of -8.0 D. 
We will predict both AL value and myopia status (the R-C task) and predict SE and AL values (the R-R task). 

\begin{figure}[!htb]
    \centering
    \includegraphics[scale = .08]{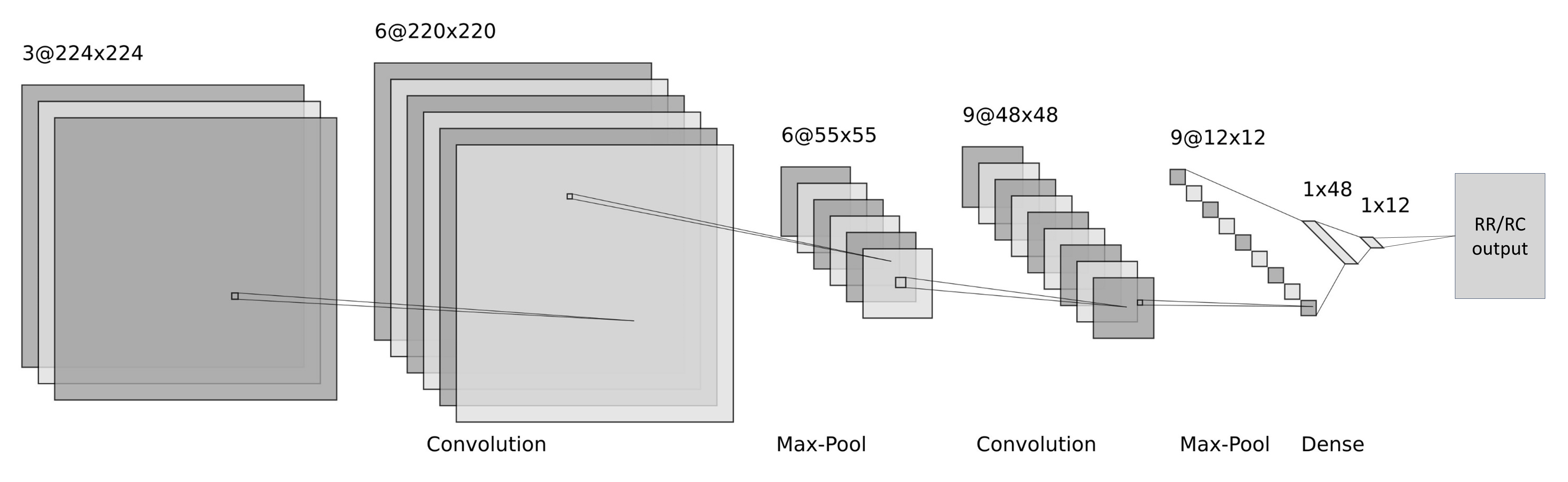}

    \includegraphics[scale = .1]{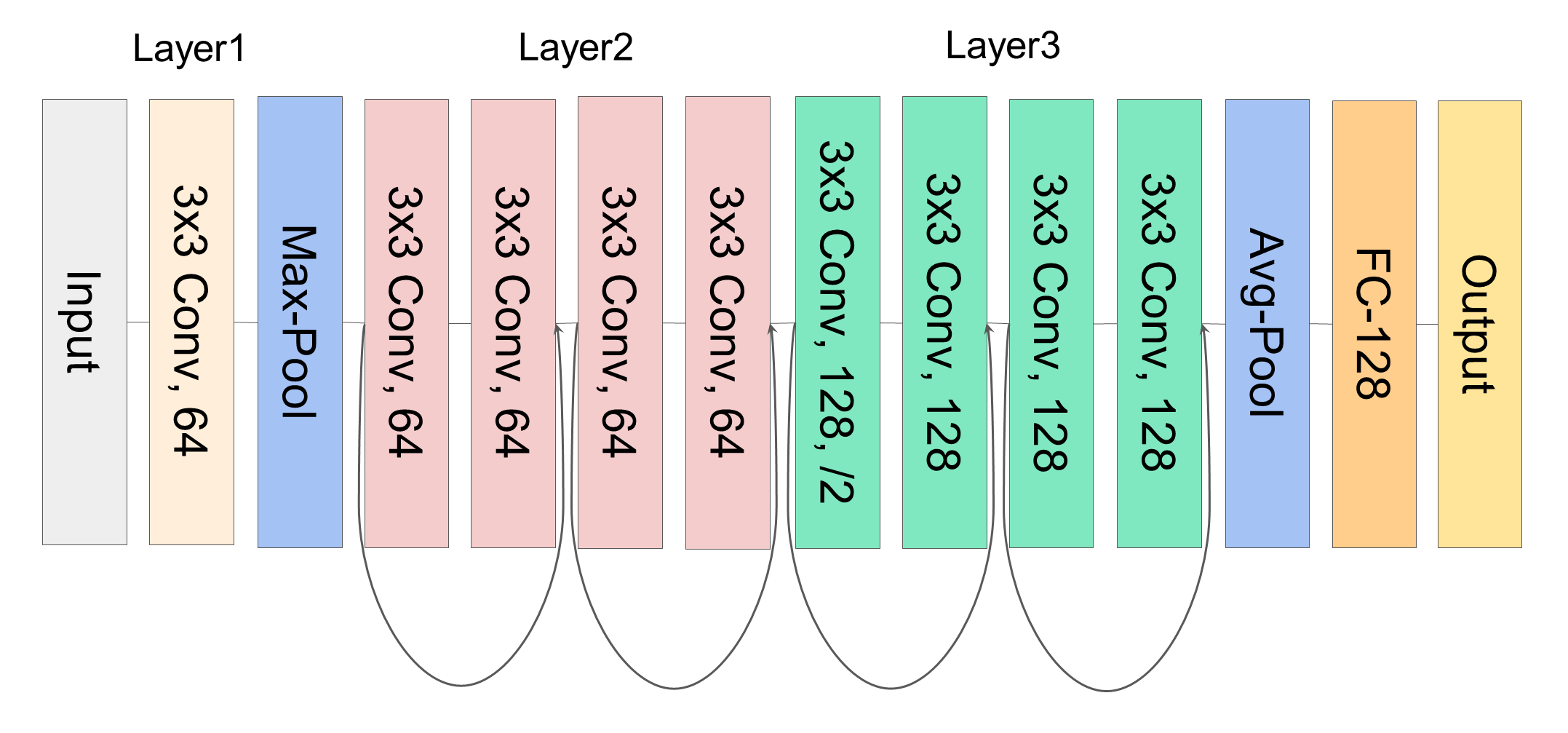}

    \includegraphics[scale = .15]{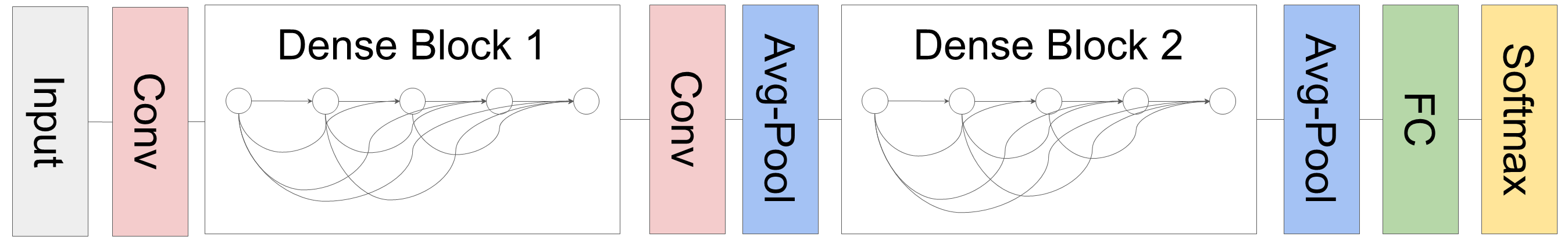}
    \caption{The architectures of backbone CNNs used. Top: architecture of simplified LeNet; middle: architecture of simplified ResNet; Bottom: architecture of simplified DenseNet. }
    \label{fig:backbonearc}
\end{figure}

{
\subsection{Structure of backbone CNNs}\label{subsec:structure_CeCNN}
To validate the generality of CeCNN, we select LeNet, ResNet-18, and DenseNet as our backbone CNNs.
LeNet represents one of the simplest, original CNNs, while ResNet and DenseNet are acknowledged as two of the most effective and prominent CNN models in the field of computer vision. \\

\noindent{\textbf{Simplify backbone CNN to avoid overfitting}}~
The conventional ResNet-18 and DenseNet models contains over ten million parameters, and the original LeNet model has over five million parameters. 
However, the size of our UWF dataset is quite limited, comprising only a few hundred images. 
This discrepancy posed a significant challenge in the form of severe overfitting, that results in  a large gap between the losses on the training set and the validation set \citep{goodfellow2016deep}. 
As a result, the predictive performances on our UWF test set are unsatisfactory, no matter what kind of loss is used to train the backbone CNN. 

To mitigate the overfitting issue caused by the overparameterized backbone CNNs, we adopt the common strategy to reduce the number of the learning parameters by simplifying the neural network architecture (\cite{han2015learning}; \cite{keshari2018learning}; among others). 
Specifically, we removed the last two CNN blocks from ResNet18, removed the last dense block from DenseNet, and increased the filter sizes of the convolutional and pooling layers of LeNet. 
This substantial reduction in the parameter count effectively prevented overfitting. After simplification, for example, the parameter count of ResNet18 reduced from over ten million to around six hundred thousand. 
Figure \ref{fig:backbonearc} shows the architectures of the simplified versions of LeNet, ResNet18 and DenseNet. 
These simplified backbone CNNs are also set as the baselines for comparison. 

Simplifying the backbone CNNs on our UWF dataset does NOT indicate that CeCNN has limitations in the choice of backbone CNN. 
In practical applications, if the dataset size is sufficient or if measures to address overfitting are implemented, it would not be necessary to use a simplified backbone CNN.\\
}

\begin{figure}[tb]
    \centering
    \includegraphics[scale = 0.15]{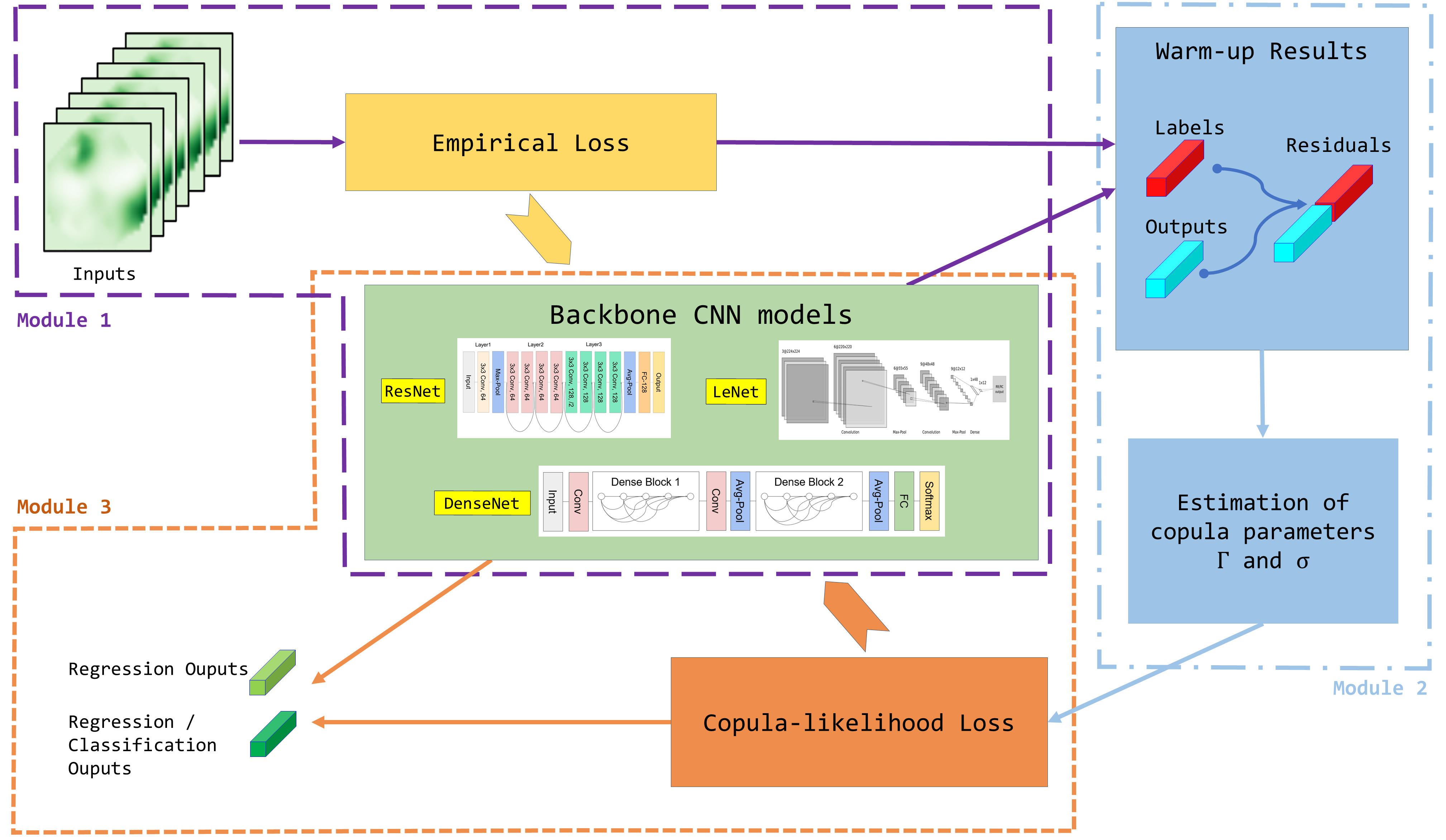}
    \caption{The whole architecture of the CeCNN framework.} 
    \label{fig:diagram_CECNN}
\end{figure}

\noindent{\textbf{Architecture of the CeCNN}}
The complete architecture for applying the CeCNN model to the UWF dataset is illustrated in Figure \ref{fig:diagram_CECNN}. 
We begin with a warm-up CNN module where we train the backbone CNNs under empirical loss (Module 1). Then, we estimate the copula parameter based on the residuals and Gaussian scores obtained in the warm-up CNN (Module 2). 
Finally, with the estimated copula parameter, we proceed to train our CeCNN models with the proposed copula-likelihood loss (Module 3). 
\\

{
\subsection{Results on the UWF dataset}
To evaluate the predictive capability of CeCNN, we consider two competitors for comparison. 
The first competitor is the so-called \textit{baseline}, where the backbone CNN is equipped with the empirical losses for the R-C and R-R tasks, respectively. 
The other competitor is the uncertainty loss \citep{kendall2018multi}, 
a commonly used multi-task learning loss in the DL community. 
Comparison with the uncertainty loss is deferred to Supplement B. 
We do not compare with the Pareto weighted loss \citep{lin2019pareto} due to the high computational burden in tuning the weights.

We partition the full UWF dataset into the training data set, the validation set, and the testing set with a ratio of 6:2:2. 
In the R-C tasks, we evaluate predictive performance by classification accuracy and AUC for the classification task, and the  Root Mean Square Error (RMSE) for the regression task; 
in the R-R tasks, we use the RMSE and the Mean Absolute Error (MAE) as metrics. \\
}

\begin{figure}[!htb]
    \centering
    \includegraphics[scale = .3]{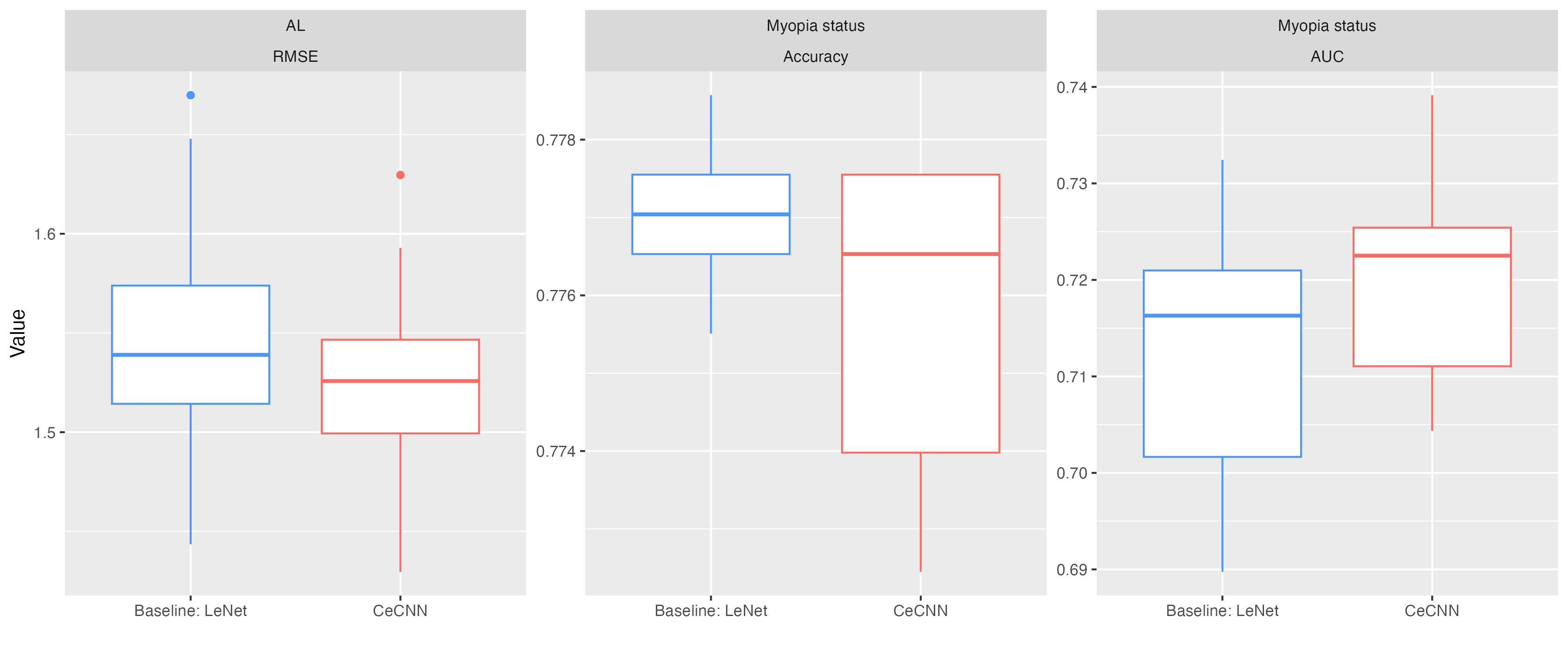} 

    \includegraphics[scale = .3]{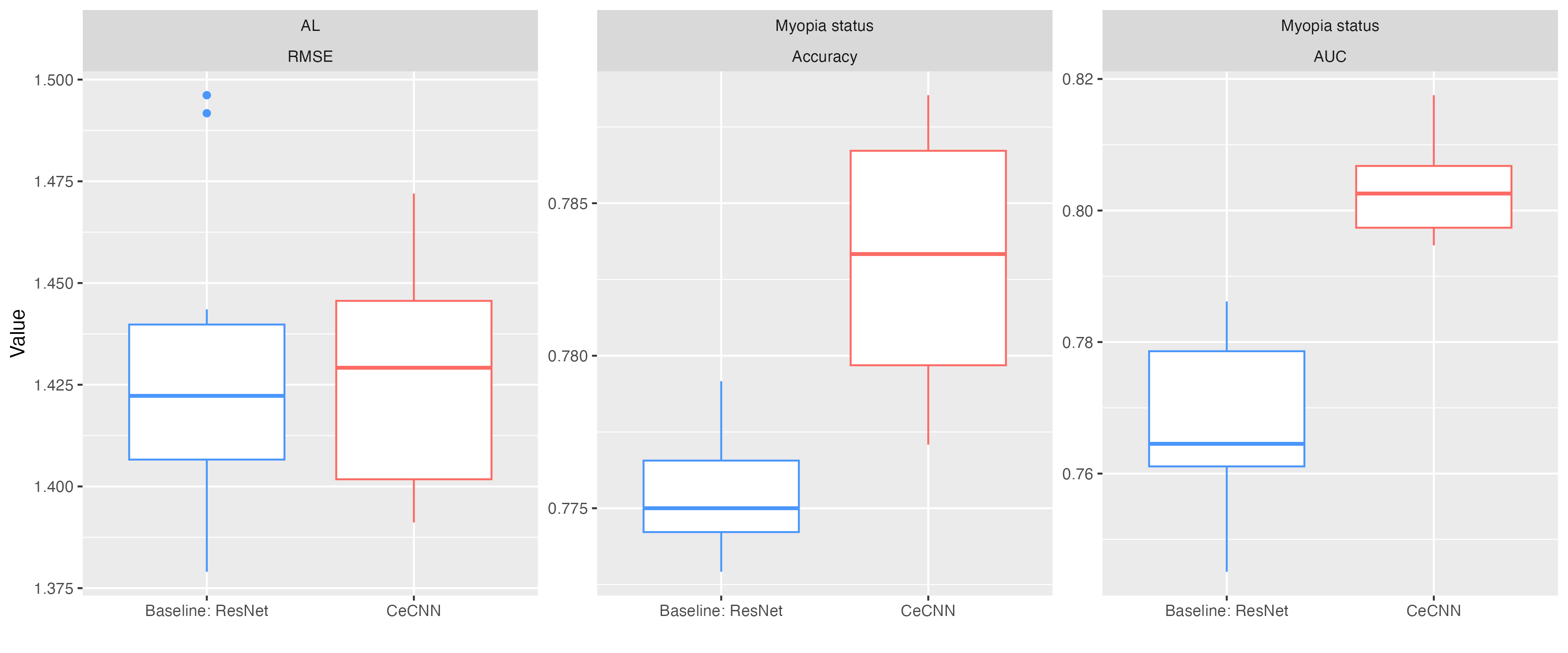}

\includegraphics[scale = .3]{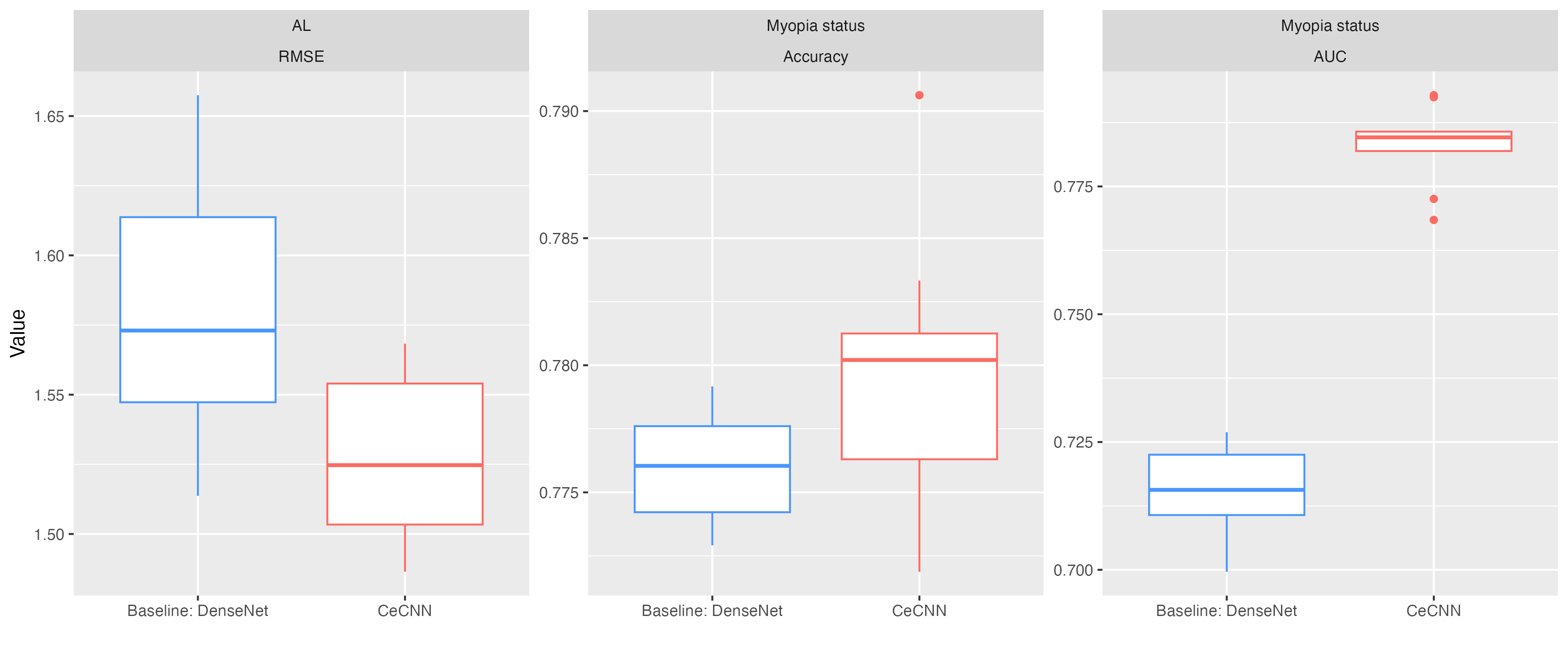}
    \caption{Box plots of RMSE, classficiation accuracy and AUC in 10 rounds of 5-fold validation of the R-C tasks. Top: with LeNet backbone; middle: with ResNet backbone; bottom: with DenseNet backbone. 
    } 
    \label{fig: res_R-C}
\end{figure}

{
\noindent{\textbf{Results on the R-C task}}~
We present the boxplots of the evaluation metrics for the R-C tasks in Figure \ref{fig: res_R-C}. 
{
Compared with the baseline, with a significance level of 0.05, the CeCNN significantly improves the AUC with both the DenseNet and the ResNet backbones. 
Furthermore, the CeCNN with the DenseNet backbone absolutely reduces the average regression RMSE by 2.887\% and slightly improves the classification accuracy by 0.516\%; 
the CeCNN  with the ResNet backbone slightly reduces the RMSE by 0.120\% and absolutely increases the classification accuracy by 0.994\%; 
CeCNN with the LeNet backbone reduces the RMSE by 1.223\%, and increases the AUC by 1.227\%, while slightly sacrificing classification accuracy by 0.145\%. 
In summary, the CeCNN enhances the baseline in almost all tasks with various backbone CNNs; the enhancement on the DenseNet and the ResNet backbones is more significant than that on the LeNet backbone. 
We conjecture that this phenomenon may be caused by the different width of the last F-C layer of different CNNs, that is, the number $K$ in Theorem \ref{theo: lower variance}. 
Recall that Assumption \ref{ass: uncover} in Section \ref{sec: rethinking} requires the activated feature maps for the two tasks to be uncovered.
When $K$ is relatively small (e.g. $K=12$ in the LeNet), this assumption may be violated, limiting the possible enhancement by the copula-likelihood loss. 
In contrast, for ResNet and DenseNet where $K$, the input dimension of the last F-C layer is large, the enhancement from using the copula-likelihood loss is absolute. }
\\
}

\begin{figure}[tb]
    \centering
    \includegraphics[scale = .3]{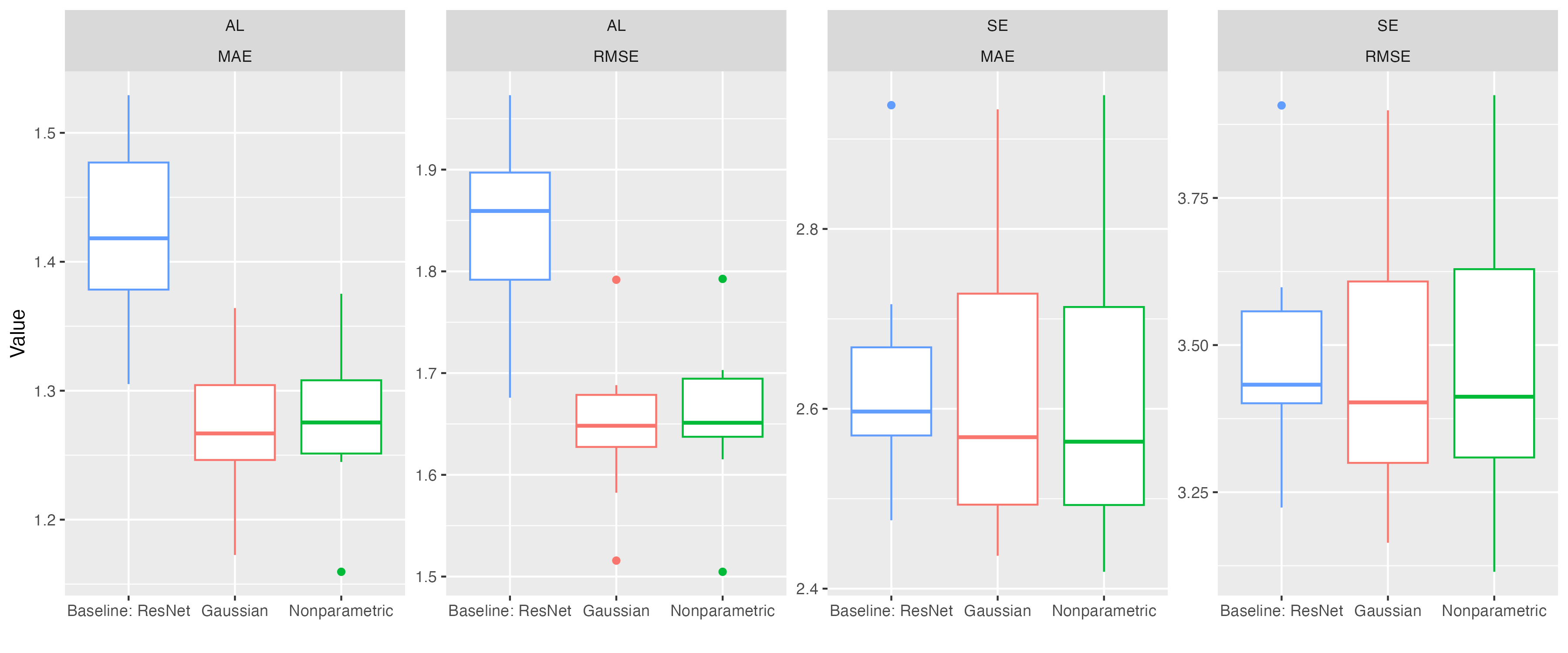}

      \includegraphics[scale = .3]{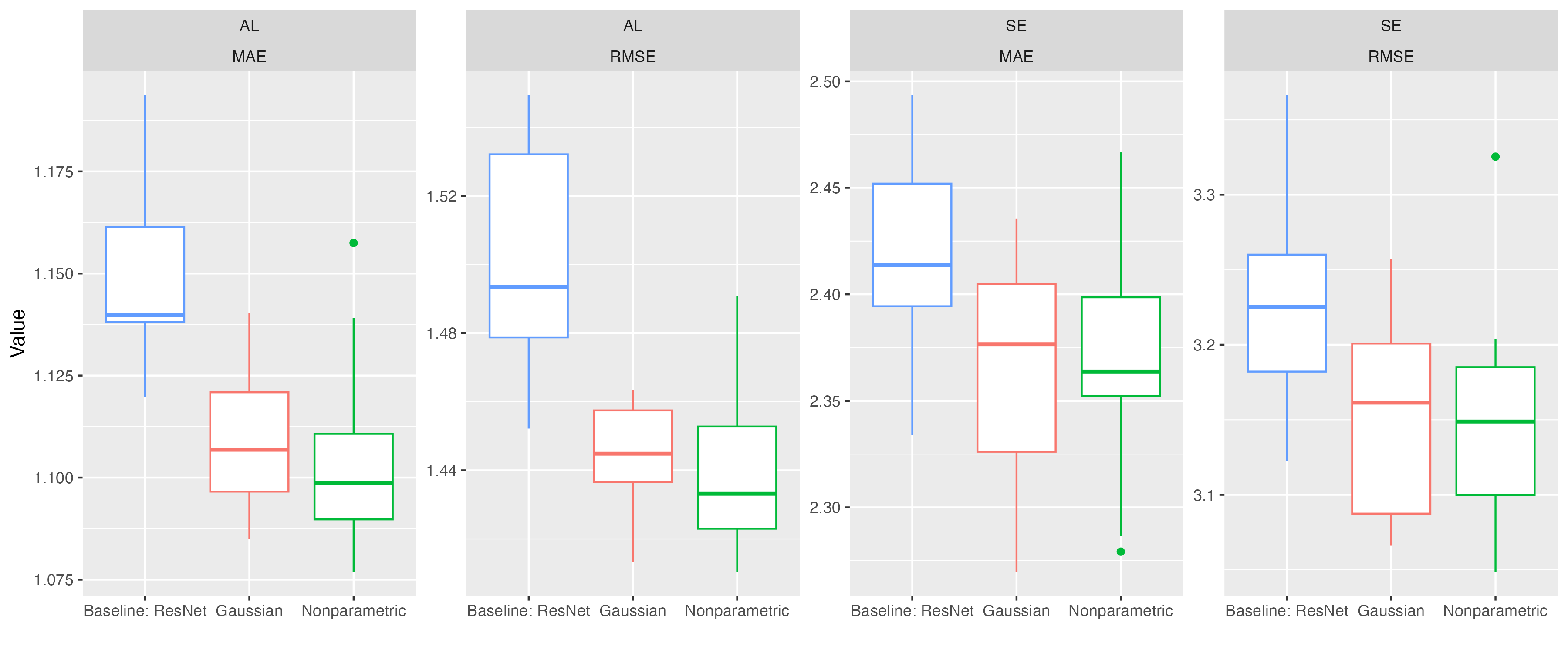}
      
     \includegraphics[scale = .3]{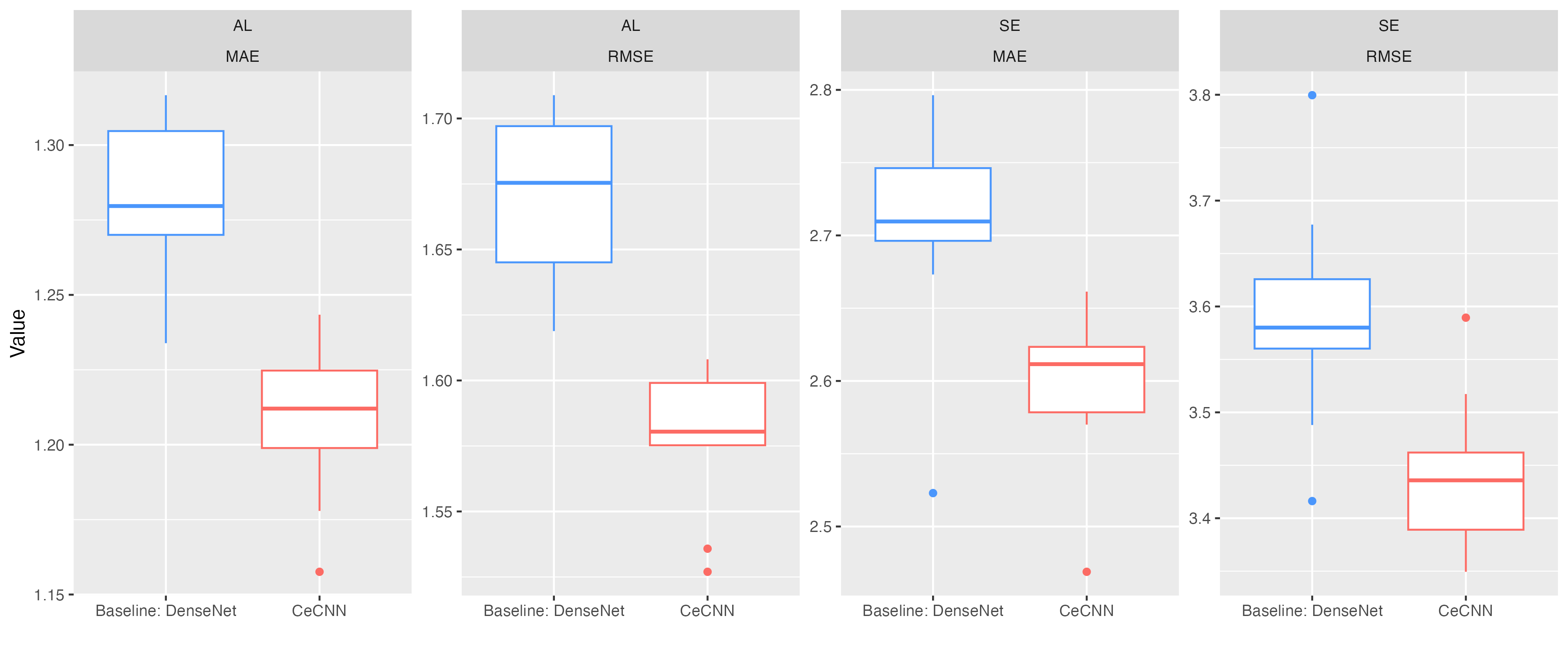}
      
    \caption{Box plots of RMSE and MAE in 10 rounds of 5-fold validation of the R-R tasks. Top: with the LeNet backbone; middle: with the ResNet backbone; bottom: with the DenseNet backbone. } 
    \label{fig: res_R-R}
\end{figure}

{
\noindent{\textbf{Results on the R-R task}} ~ We present boxplots of the evaluation metrics for the R-C tasks in Figure \ref{fig: res_R-R}. 
On average,  when using the Gaussian error, CeCNN reduces the RMSE of SE by 0.717\%, 2.232\% , and 3.827\%, and
reduces the RMSE of AL by 10.378\%, 3.683\%, and 5.262\%, on the LeNet, the ResNet, and the DenseNet backbones, respectively. 
{
With a significance level of 0.05, the improvements in the AL prediction with all of the LeNet, the ResNet, and the DenseNet backbones are significant, and the improvement in the SE prediction  with the DenseNet backbone is also significant. }
When using the nonparametric error, the predictive performances of the CeCNNs  are similar to those using Gaussian error. 
It is not surprising that the CeCNN enhances predicting AL more than it enhances predicting SE since the marginal variance of AL is smaller than that of SE. 
In the joint Gaussian log-likelihood \eqref{likelihood}, a smaller variance puts a larger weight on the corresponding loss, and hence the optimizer tends to optimize more on AL than SE. 
}

\section{Simulations on synthetic data}
\label{sec:sim}
We carry out simulations on synthetic datasets for both R-C and R-R tasks to illustrate the superiority of CeCNN compared with the baseline. 
To match our application scenarios, in both R-C and R-R tasks, we consider correlated bivariate responses and a single image covariate. 
We present the simulation details for the regression-classification and regression-regression tasks in subsections  \ref{subsec:sim_reg-class} and \ref{subsec:sim_reg-reg}, respectively.

\subsection{Simulated regression-classification task}
\label{subsec:sim_reg-class}

On this task, the synthetic image covariate is generated as a grey matrix $X_i \in \mathbb{R}^{9 \times 9}$ that can be divided into nine independent blocks of $3 \times 3$ square block matrices $X_{t, s} \in \mathbb{R}^{3\times 3}$:
\begin{equation*}
X = 
\begin{pmatrix}
X_{1,1} & X_{1,2} & X_{1,3} \\
X_{2,1} & X_{2,2} & X_{2,3} \\
X_{3,1} & X_{3,2} & X_{3,3}
\end{pmatrix}, 
~X_{t, s} = 
\begin{pmatrix}
X_{t,s}^{(1, 1)} & X_{t,s}^{(1, 2)} & X_{t,s}^{(1, 3)} \\
X_{t,s}^{(2, 1)} & X_{t,s}^{(2, 2)} & X_{t,s}^{(2, 3)} \\
X_{t,s}^{(3, 1)} &X_{t,s}^{(3, 2)} &X_{t,s}^{(3, 3)}
\end{pmatrix}, 
~t, s =1, 2, 3. 
\end{equation*}
For $k, l = 1, 2, 3$, we set $X_{3, 3}^{(k, l)} \sim N(1, 0.5^2)$ as elements of the dark block and $X_{t, s}^{(k, l)} \sim N(0, 1^2)$ as elements of bright blocks, for $(t, s) \not = (3, 3)$. 
Based on this covariate, we generated the synthetic continuous response $y_1 \in \mathbb{R}$ and the synthetic binary response $y_2 \in \{0, 1\}$ using model \eqref{mod:basic}. 

For each block $X_{t, s}$, we define the block operators $S_{t, s}, S^*_{t, s}: \mathbb{R}^{3\times 3} \to \mathbb{R}$ for responses $y_1$ and $y_2$, respectively. 
Then the two responses were generated as 
\begin{eqnarray}
\label{sim:RC}
    \begin{aligned}
        y_1 \sim N\left\{\sum_{1\le t, s\le 3}S_{t, s} (X_{t, s}), 1^2\right\}, ~
        y_2 \sim \text{Bernoulli}\left\{\mathcal{S}[\sum_{1\le t, s\le 3}S^*_{t, s} (X_{t, s})]\right\}. 
    \end{aligned}
\end{eqnarray}
For the block operators $S_{t, s}$, we set 
\begin{eqnarray*}
    \begin{aligned}
  S_{1, 1} = \sum_{1\le l, k \le 3} \text{tanh}(X_{1, 1}^{(k, l)}), ~ S_{2, 2} =  \sum_{1\le l, k \le 3}X_{2, 2}^{(k, l)}, ~ S_{3, 3} = \text{tanh}\left( \sum_{1\le l, k \le 3}X_{3, 3}^{(k, l)}\right),
    \end{aligned}
\end{eqnarray*}
and $S_{t, s} = 0$ for other blocks. 
Under this setting, the function $g_1 = \sum_{1 \le t, 2, \le 3}S_{t, s}$ is a nonlinear function that can be modeled by a CNN. 
For the block operators $S^*_{t, s}$, we set 
$$
S^*_{2, 2} = S_{2, 2} =  \sum_{1\le l, k \le 3}X_{2, 2}^{(k, l)}
$$
and $S^*_{t, s} = 0$ for other blocks so that $E(y_2) = 1/2$.
The two responses are dependent in this case since they share the same block operator on block $X_{2, 2}$.

\subsection{Simulated regression-regression task}\label{subsec:sim_reg-reg}
For the regression-regression task, we generate the synthetic image covariates in a similar way to that of the regression-classification task. 
For this task, we set five dark blocks $X_{t, s}^{(k,l)} \sim N(1, 0.5^2)$ for $k, l = 1, 2, 3$ and $(t, s) \in \{(1, 1), (2, 2), (3, 3), (1, 3), (3, 1)\}$.
The remaining blocks are set to be the aforementioned bright blocks $N(0, 1^2)$. 
The responses are generated from model \eqref{pureres}, where the model error $\bm{\epsilon}_i = (\epsilon_{i1}, \epsilon_{i2})\sim MVN(\bm{0}_2, \Sigma)$ with 
\begin{equation*}
\begin{aligned}
\Sigma = 
\begin{pmatrix}
1 & 0  \\
0 & 2 
\end{pmatrix}
\begin{pmatrix}
1 & 0.7  \\
0.7 & 1 
\end{pmatrix}
\begin{pmatrix}
1 & 0  \\
0 & 2
\end{pmatrix}. 
\end{aligned}
\end{equation*}
This covariance setting yields a strong correlation and two different levels of marginal variation (variances of 1 and 4), corresponding to the AL and SE, respectively. 
We set the nonlinear regression functions $g_1$ and $g_2$ as the sum of block operators so that 
\begin{align*}
 g_1(X) = \sum_{1\le t, s \le 3} S_{t, s} (X_{t,s}), ~  g_2(X) = \sum_{1\le t, s \le 3} S_{t, s}^* (X_{t,s}). 
\end{align*}
For $y_1$, the block operators $S_{t, s}$ are given by 
$$
    S_{1, 1} = \sum_{1\le l, k \le 3} \text{tanh}(X_{1, 1}^{(k, l)}), ~ S_{2, 2} =  \sum_{1\le l, k \le 3}X_{2, 2}^{(k, l)}, ~ S_{3, 3} = \text{tanh}\left( \sum_{1\le l, k \le 3}X_{3, 3}^{(k, l)}\right), 
$$
and $S_{t, s} = 0$ for other pairs of $(t, s)$. 
For $y_2$, the the block operators $S_{t, s}^*$ are given by 
\begin{align*}
    S_{1, 3}^* = \sum_{1\le l, k \le 3} \text{tanh}(X_{1, 3}^{(k, l)}), ~ S_{2, 2}^* =  \sum_{1\le l, k \le 3}X_{2, 2}^{(k, l)}, ~ S_{3, 3}^* = \text{tanh}\left( \sum_{1\le l, k \le 3}X_{3, 3}^{(k, l)}\right),
\end{align*}
and $S_{t, s}^* = 0$ for other pairs of $(t, s)$. 
Under this setting, marginally we have 
$
E(y_1) = E(y_2). 
$
{
Note that the block operators $S_{t, x}$ and $S_{t, s}^*$ can be viewed as the input feature maps of the last F-C layer of the CNN. 
Then our setting guarantees that Assumption \ref{ass: uncover} holds. }

\subsection{Simulation results}\label{subsec:sim_model}
\begin{figure}[tb]
    \centering
    \includegraphics[scale = .3]{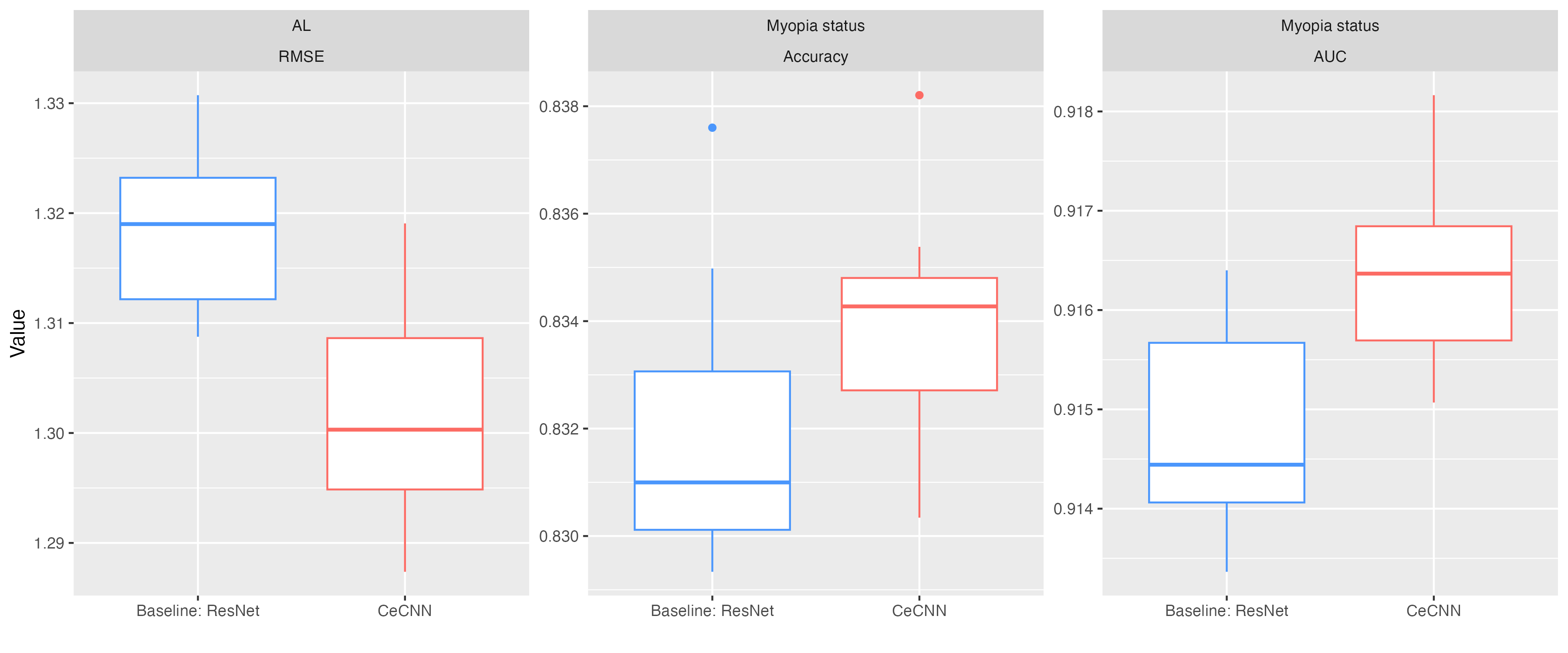}

    \includegraphics[scale = .3]{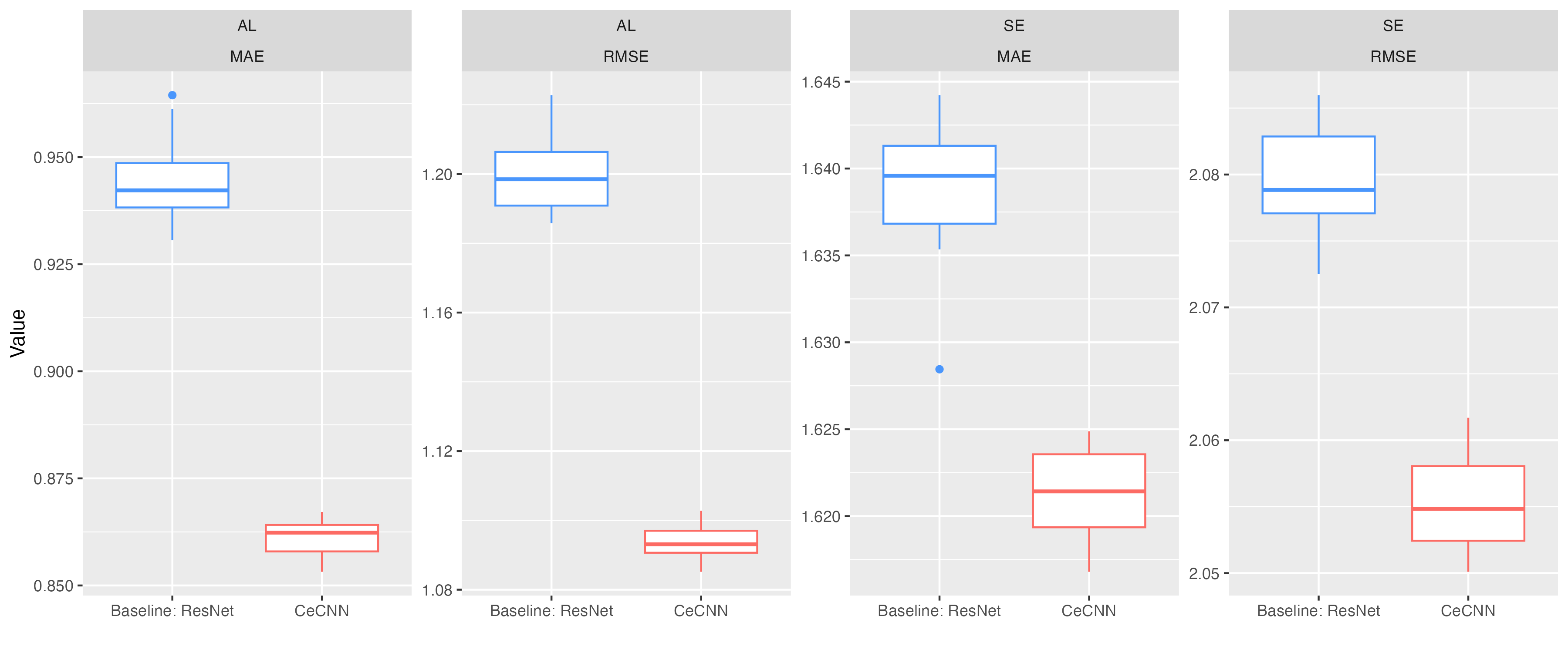}
    \caption{Top: box plots of RMSE, classification accuracy and AUC in 10 rounds of 5-fold validation of the R-C tasks on the synthetic dataset; bottom: box plots of RMSE and MAE for two responses in 10 rounds of 5-fold validation of the R-R tasks on the synthetic dataset. } 
    \label{fig:res_simulation}
\end{figure}

{In total, we generated $n=10,000$ synthetic images and the corresponding pairs of responses in our simulations.
To save computation time, we adopt a simple backbone CNN in both the baseline and the CeCNN. 
The backbone model used in the simulations consists of two basic convolutional layers and two fully connected layers. 
With such a simple backbone CNN and the relatively large data size, the overfitting issue is not severe in our simulations. 
The baseline is then defined as the same backbone CNN equipped with the empirical loss. 
For the R-R task, we used MSE, AUC, and classification accuracy as the evaluation metrics; 
for the R-R tasks, we use RMSE as the evaluation metric. 

From Figure \ref{fig:res_simulation}, we find that CeCNN improves the prediction performance of the baseline in both R-C and R-R tasks. 
Recall that Assumption \ref{ass: uncover} holds under our R-C simulation setting. 
The simulation result on the R-R task illustrates the improvement brought by CeCNN that we established theoretically in Theorem \ref{theo: lower variance}. 
}

\section{Discussion}
\label{sec:discussion}
{
In this paper, we propose the CeCNN, a new one-stop ophthalmic AI framework for myopia screening based on the UWF fundus image.
The CeCNN models and incorporates the conditional dependence among mixed-type responses in a multiresponse nonlinear regression through a new copula-likelihood loss to train the backbone CNN. 
We provide explicit statistical interpretations of the conditional dependence and justify the enhancement in estimation brought about by the proposed loss through a heuristic inferential procedure. 
In our UWF dataset, the CeCNN succeeds in enhancing the predictive capability of  various deep learning models in measuring SE, AL, and diagnosing high myopia simultaneously.



\subsubsection*{Why not iteratively update the copula parameters} 
One referee questioned why we do not iteratively update the estimate of the copula parameter $\Gamma$ and $\sigma$ in module 2, during each epoch updating of the backbone CNN in module 3 of the CeCNN. 
Indeed, we estimate the copula parameter $\Gamma$ from the outputs of the backbone CNN trained under the empirical loss. 
It is called the warm-up CNN since we only train this CNN once to estimate $\Gamma$ and $\sigma$. 
We attempt to answer the reviewer's question from the point of view of estimating $\bm{w}_j$, the weights in the last F-C layer of the backbone CNN. 

In the CeCNN, by replacing the true $\Gamma$ with its estimate based on the residuals and Gaussian scores obtained from the warm-up CNN, the induced estimator $\hat{\bm{w}}_j^{cop}$ is the feasible generalized least square (FGLS) estimator \citep{avery1977error}. 
Since the FGLS estimator is asymptotically equivalent to the GLS estimator using the true copula parameters \citep{prucha1984asymptotic},  using the estimated copula parameters in the copula-likelihood loss will not lose estimation efficiency asymptotically. 
This justifies our CeCNN structure and the utility of the first two modules. 

In contrast, iteratively updating $\Gamma$ and $\sigma$ may not be stable. 
The estimator $\hat{\bm{w}}_j^{cop}$ obtained using the iteratively updated copula parameter is closely related to the quadratic maximum likelihood estimator \citep{moon2006seemingly} obtained by alternating minimization \citep{jain2015alternating}. 
In traditional alternating minimization, in each loop of the copula parameter, one has to optimize the copula-likelihood loss within one epoch of iterations. 
Unfortunately, it is almost impossible for the optimizer to converge within one epoch in deep learning. 
Consequently, the updated copula parameter in each epoch is unstable in the sense that there is no guarantee that the updated copula parameter decreases the copula-likelihood loss. 
As a result, we find that iteratively updating the copula parameter leads to unsatisfactory predictive performances; refer to Supplement C for details. 

\subsubsection*{Why only reduce the model size to avoid overfitting}
Another referee questioned why we do not consider other techniques to overcome overfitting, such as hyperparameter tuning, regularization, and data augmentation. 
We attempt to justify this point here. 

The CeCNN framework realies on a backbone CNN. 
In both the warm-up CNN and the C-CNN modules, we maintain the default settings for most of the hyperparameters in the backbone CNN, allowing ophthalmic practitioners to use the CeCNN conveniently without further tuning. 
We carefully tune the learning rate on the warm-up CNN such that the backbone CNN converges under the training loss and no longer improves in the validation loss. 
Then in the C-CNN module, we reduce the learning rate of the warm-up module by multiplying by $0.1$ to seek a more precise direction for optimization. 
We do not consider dropout rate tuning of the backbone CNN since, unfortunately, we found that dropout leads to poor prediction performance in both the R-C and the R-R tasks. 
We conjecture the reason is that dropout cannot preserve the variance in each hidden layer after the nonlinear activation and thus, may not be suitable for regression tasks \citep{ozgur2020effect}.

Adding regularization to the CNN weights is generally thought to be a good choice to overcome overfitting. 
However, in practice, it is challenging to specify the tuning parameter $\lambda$ for the regularization. 
In our practice, we try several candidate values for the tuning parameters and present two examples of the loss traces in Supplement D. 
We find that on our UWF dataset, with a small $\lambda$, the overfitting issue is still serious. 
Otherwise, with a large $\lambda$, the predictive performance on the test set is worse than that with no regularization. 
Therefore, we do not consider regularization in our application to our UWF dataset. 

Data augmentation is effective in improving DL models with limited data size. 
There are a wealth of approaches to data augmentation in computer vision including image manipulation, image eraser, and image mix, among others. 
We believe that suitable data augmentation can lead to better predictive performance of both the baseline and the CeCNN. 
Nonetheless, it is challenging to find ``suitable" data augmentation methods since generalizing qualified synthetic images based on the very limited data size is difficult \citep{yang2022image}. 
To avoid possible error induced by poorly generated images, we do not consider data augmentation procedures in our application. 
Seeking an appropriate data augmentation approach for our UWF dataset is beyond the scope of the present study but it deserves treating in separate work.



{
\subsubsection*{Future work}
The present paper has shown that the proposed copula-likelihood loss can reduce the asymptotic estimation risk in the R-R task, while theoretically proving the enhancement in the R-C task is much more difficult , pending another separate work to resolve. 
From the application perspective, this paper only considers bivariate myopia screening tasks on a single eye. 
The proposed loss was successfully applied to a quadravariate case where the images of left and right eyes are included to predict SE and AL for both eyes \citep{li2024oucopula}. 
It is anticipated that besides myopia screening in ophthalmology, CeCNN can also be applied to other multi-task learning scenarios. 
Furthermore, an interesting future direction is to associate the CeCNN framework with large models such as Vision Transformers \citep{dosovitskiy2020image}. 
This can be accomplished by replacing the backbone CNNs by transformers if the overfitting issues are well-resolved. }

\begin{funding}
Chong Zhong is partially supported by the ZZPC, PolyU and Postdoc Fellowship of CAS AMSS-PolyU Joint Laboratory of Applied Mathematics. 
Bo Fu and Yang Li are partially supported by the National Natural Science Foundation of China (Grant No. 71991471).
Danjuan Yang and Meiyan Li are partially supported by the Shanghai Rising-Star Program (21QA1401500). 
Meiyan Li is also supported by the National Natural Science Foundation of China (82371091).
Catherine C. Liu is partially supported by a grant (GRF15301123) from the Research Grants Council of the Hong Kong SAR. 
A. H. Welsh is partially supported by the Australian Research Council Discovery Project DP230101908.
\end{funding}

\bibliographystyle{apalike}
\bibliography{Ref}

\begin{thebibliography}{}

\bibitem[Avery, 1977]{avery1977error}
Avery, R.~B. (1977).
\newblock Error components and seemingly unrelated regressions.
\newblock {\em Econometrica: Journal of the Econometric Society}, pages 199--209.

\bibitem[Cen et~al., 2021]{cen2021automatic}
Cen, L.-P., Ji, J., Lin, J.-W., Ju, S.-T., Lin, H.-J., Li, T.-P., Wang, Y., Yang, J.-F., Liu, Y.-F., Tan, S., et~al. (2021).
\newblock Automatic detection of 39 fundus diseases and conditions in retinal photographs using deep neural networks.
\newblock {\em Nature communications}, 12(1):4828.

\bibitem[Chen et~al., 2019a]{chen2019non}
Chen, H., Raskutti, G., and Yuan, M. (2019a).
\newblock Non-convex projected gradient descent for generalized low-rank tensor regression.
\newblock {\em Journal of Machine Learning Research}, 20(5):1--37.

\bibitem[Chen et~al., 2019b]{chen2019multi}
Chen, Z.-M., Wei, X.-S., Wang, P., and Guo, Y. (2019b).
\newblock Multi-label image recognition with graph convolutional networks.
\newblock In {\em Proceedings of the IEEE/CVF conference on computer vision and pattern recognition}, pages 5177--5186.

\bibitem[Dai et~al., 2022]{dai2022significance}
Dai, B., Shen, X., and Pan, W. (2022).
\newblock Significance tests of feature relevance for a black-box learner.
\newblock {\em IEEE transactions on Neural Networks and Learning Systems}, 35(2):1898--1911.

\bibitem[De'Ath, 2002]{de2002multivariate}
De'Ath, G. (2002).
\newblock Multivariate regression trees: a new technique for modeling species--environment relationships.
\newblock {\em Ecology}, 83(4):1105--1117.

\bibitem[Dosovitskiy et~al., 2020]{dosovitskiy2020image}
Dosovitskiy, A., Beyer, L., Kolesnikov, A., Weissenborn, D., Zhai, X., Unterthiner, T., Dehghani, M., Minderer, M., Heigold, G., Gelly, S., et~al. (2020).
\newblock An image is worth 16x16 words: Transformers for image recognition at scale.
\newblock In {\em International Conference on Learning Representations}.

\bibitem[Goodfellow et~al., 2016]{goodfellow2016deep}
Goodfellow, I., Bengio, Y., and Courville, A. (2016).
\newblock {\em Deep learning}.
\newblock MIT press.

\bibitem[Haarman et~al., 2020]{haarman2020complications}
Haarman, A.~E., Enthoven, C.~A., Tideman, J. W.~L., Tedja, M.~S., Verhoeven, V.~J., and Klaver, C.~C. (2020).
\newblock The complications of myopia: a review and meta-analysis.
\newblock {\em Investigative ophthalmology \& visual science}, 61(4):49--49.

\bibitem[Han et~al., 2015]{han2015learning}
Han, S., Pool, J., Tran, J., and Dally, W. (2015).
\newblock Learning both weights and connections for efficient neural network.
\newblock {\em Advances in neural information processing systems}, 28.

\bibitem[He et~al., 2016]{he2016deep}
He, K., Zhang, X., Ren, S., and Sun, J. (2016).
\newblock Deep residual learning for image recognition.
\newblock In {\em Proceedings of the IEEE conference on computer vision and pattern recognition}, pages 770--778.

\bibitem[Huang et~al., 2017]{huang2017densely}
Huang, G., Liu, Z., Van Der~Maaten, L., and Weinberger, K.~Q. (2017).
\newblock Densely connected convolutional networks.
\newblock In {\em Proceedings of the IEEE conference on computer vision and pattern recognition}, pages 4700--4708.

\bibitem[Iwase et~al., 2006]{iwase2006prevalence}
Iwase, A., Araie, M., Tomidokoro, A., Yamamoto, T., Shimizu, H., Kitazawa, Y., Group, T.~S., et~al. (2006).
\newblock Prevalence and causes of low vision and blindness in a japanese adult population: the tajimi study.
\newblock {\em Ophthalmology}, 113(8):1354--1362.

\bibitem[Jain and Tewari, 2015]{jain2015alternating}
Jain, P. and Tewari, A. (2015).
\newblock Alternating minimization for regression problems with vector-valued outputs.
\newblock {\em Advances in Neural Information Processing Systems}, 28.

\bibitem[Kendall et~al., 2018]{kendall2018multi}
Kendall, A., Gal, Y., and Cipolla, R. (2018).
\newblock Multi-task learning using uncertainty to weigh losses for scene geometry and semantics.
\newblock In {\em Proceedings of the IEEE conference on computer vision and pattern recognition}, pages 7482--7491.

\bibitem[Keshari et~al., 2018]{keshari2018learning}
Keshari, R., Vatsa, M., Singh, R., and Noore, A. (2018).
\newblock Learning structure and strength of cnn filters for small sample size training.
\newblock In {\em proceedings of the IEEE conference on computer vision and pattern recognition}, pages 9349--9358.

\bibitem[Kim et~al., 2021]{kim2021development}
Kim, K.~M., Heo, T.-Y., Kim, A., Kim, J., Han, K.~J., Yun, J., and Min, J.~K. (2021).
\newblock Development of a fundus image-based deep learning diagnostic tool for various retinal diseases.
\newblock {\em Journal of Personalized Medicine}, 11(5):321.

\bibitem[Kingma and Ba, 2014]{kingma2014adam}
Kingma, D.~P. and Ba, J. (2014).
\newblock Adam: A method for stochastic optimization.
\newblock {\em arXiv preprint arXiv:1412.6980}.

\bibitem[Kobayashi et~al., 2005]{kobayashi2005fundus}
Kobayashi, K., Ohno-Matsui, K., Kojima, A., Shimada, N., Yasuzumi, K., Yoshida, T., Futagami, S., Tokoro, T., and Mochizuki, M. (2005).
\newblock Fundus characteristics of high myopia in children.
\newblock {\em Japanese Journal of Ophthalmology}, 49:306--311.

\bibitem[Lai et~al., 2023]{lai2023single}
Lai, Q., Zhou, J., Gan, Y., Vong, C.-M., and Chen, C.~P. (2023).
\newblock Single-stage broad multi-instance multi-label learning (bmiml) with diverse inter-correlations and its application to medical image classification.
\newblock {\em IEEE Transactions on Emerging Topics in Computational Intelligence}.

\bibitem[LeCun et~al., 2015]{lecun2015deep}
LeCun, Y., Bengio, Y., and Hinton, G. (2015).
\newblock Deep learning.
\newblock {\em Nature}, 521(7553):436--444.

\bibitem[LeCun et~al., 1998]{lecun1998gradient}
LeCun, Y., Bottou, L., Bengio, Y., and Haffner, P. (1998).
\newblock Gradient-based learning applied to document recognition.
\newblock {\em Proceedings of the IEEE}, 86(11):2278--2324.

\bibitem[Li et~al., 2024]{li2024oucopula}
Li, Y., Huang, Q., Zhong, C., Yang, D., Li, M., Welsh, A., Liu, A., Fu, B., Liu, C.~C., and Zhou, X. (2024).
\newblock {OUCopula: Bi-Channel Multi-Label Copula-Enhanced Adapter-Based CNN for Myopia Screening Based on OU-UWF Images}.
\newblock IJCAI.

\bibitem[Li et~al., 2021]{li2021deep}
Li, Z., Guo, C., Lin, D., Nie, D., Zhu, Y., Chen, C., Zhao, L., Wang, J., Zhang, X., Dongye, M., et~al. (2021).
\newblock Deep learning for automated glaucomatous optic neuropathy detection from ultra-widefield fundus images.
\newblock {\em British journal of ophthalmology}, 105(11):1548--1554.

\bibitem[Lian et~al., 2022]{lian2022multi}
Lian, C., Liu, M., Wang, L., and Shen, D. (2022).
\newblock {Multi-task weakly-supervised attention network for dementia status estimation with structural MRI}.
\newblock {\em IEEE Transactions on Neural Networks and Learning Systems}, 33(8).

\bibitem[Lin et~al., 2019]{lin2019pareto}
Lin, X., Zhen, H.-L., Li, Z., Zhang, Q.-F., and Kwong, S. (2019).
\newblock Pareto multi-task learning.
\newblock {\em Advances in neural information processing systems}, 32.

\bibitem[Liu et~al., 2019]{liu2019joint}
Liu, M., Zhang, J., Adeli, E., and Shen, D. (2019).
\newblock {Joint classification and regression via deep multi-task multi-channel learning for Alzheimer's disease diagnosis}.
\newblock {\em IEEE Transactions on Biomedical Engineering}, 66(5):1195--1206.

\bibitem[Loh and Zheng, 2013]{loh2013regression}
Loh, W.-Y. and Zheng, W. (2013).
\newblock Regression trees for longitudinal and multiresponse data.
\newblock {\em The Annals of Applied Statistics}, pages 495--522.

\bibitem[Meng et~al., 2011]{meng2011axial}
Meng, W., Butterworth, J., Malecaze, F., and Calvas, P. (2011).
\newblock Axial length of myopia: a review of current research.
\newblock {\em Ophthalmologica}, 225(3):127--134.

\bibitem[Midena et~al., 2022]{midena2022ultra}
Midena, E., Marchione, G., Di~Giorgio, S., Rotondi, G., Longhin, E., Frizziero, L., Pilotto, E., Parrozzani, R., and Midena, G. (2022).
\newblock Ultra-wide-field fundus photography compared to ophthalmoscopy in diagnosing and classifying major retinal diseases.
\newblock {\em Scientific Reports}, 12(1):19287.

\bibitem[Moon and Perron, 2006]{moon2006seemingly}
Moon, H.~R. and Perron, B. (2006).
\newblock Seemingly unrelated regressions.
\newblock {\em The New Palgrave Dictionary of Economics}, 1(9):19.

\bibitem[Mutti et~al., 2007]{mutti2007refractive}
Mutti, D.~O., Hayes, J.~R., Mitchell, G.~L., Jones, L.~A., Moeschberger, M.~L., Cotter, S.~A., Kleinstein, R.~N., Manny, R.~E., Twelker, J.~D., and Zadnik, K. (2007).
\newblock Refractive error, axial length, and relative peripheral refractive error before and after the onset of myopia.
\newblock {\em Investigative Ophthalmology \& Visual Science}, 48(6):2510--2519.

\bibitem[Oh et~al., 2023]{oh2023deep}
Oh, R., Lee, E.~K., Bae, K., Park, U.~C., Yu, H.~G., and Yoon, C.~K. (2023).
\newblock Deep learning-based prediction of axial length using ultra-widefield fundus photography.
\newblock {\em Korean journal of ophthalmology: KJO}, 37(2):95.

\bibitem[{\"O}zg{\"u}r and Nar, 2020]{ozgur2020effect}
{\"O}zg{\"u}r, A. and Nar, F. (2020).
\newblock Effect of dropout layer on classical regression problems.
\newblock In {\em 2020 28th Signal Processing and Communications Applications Conference (SIU)}, pages 1--4. IEEE.

\bibitem[Panagiotelis et~al., 2012]{panagiotelis2012pair}
Panagiotelis, A., Czado, C., and Joe, H. (2012).
\newblock Pair copula constructions for multivariate discrete data.
\newblock {\em Journal of the American Statistical Association}, 107(499):1063--1072.

\bibitem[Prucha, 1984]{prucha1984asymptotic}
Prucha, I.~R. (1984).
\newblock On the asymptotic efficiency of feasible aitken estimators for seemingly unrelated regression models with error components.
\newblock {\em Econometrica: Journal of the Econometric Society}, pages 203--207.

\bibitem[Rahman et~al., 2017]{rahman2017integratedmrf}
Rahman, R., Otridge, J., and Pal, R. (2017).
\newblock Integratedmrf: random forest-based framework for integrating prediction from different data types.
\newblock {\em Bioinformatics}, 33(9):1407--1410.

\bibitem[Raskutti et~al., 2019]{raskutti2019convex}
Raskutti, G., Yuan, M., and Chen, H. (2019).
\newblock Convex regularization for high-dimensional multiresponse tensor regression.
\newblock {\em The Annals of Statistics}, 47(3):1554--1584.

\bibitem[Selvaraju et~al., 2017]{selvaraju2017grad}
Selvaraju, R.~R., Cogswell, M., Das, A., Vedantam, R., Parikh, D., and Batra, D. (2017).
\newblock Grad-cam: Visual explanations from deep networks via gradient-based localization.
\newblock In {\em Proceedings of the IEEE international conference on computer vision}, pages 618--626.

\bibitem[Sklar, 1959]{sklar1959fonctions}
Sklar, M. (1959).
\newblock Fonctions de repartition an dimensions et leurs marges.
\newblock {\em Publ. inst. statist. univ. Paris}, 8:229--231.

\bibitem[Song et~al., 2018]{song2018deep}
Song, L., Liu, J., Qian, B., Sun, M., Yang, K., Sun, M., and Abbas, S. (2018).
\newblock A deep multi-modal cnn for multi-instance multi-label image classification.
\newblock {\em IEEE Transactions on Image Processing}, 27(12):6025--6038.

\bibitem[Song, 2007]{peter2007correlated}
Song, P. X.-K. (2007).
\newblock {\em Correlated data analysis: modeling, analytics, and applications}.
\newblock Springer.

\bibitem[Song et~al., 2009]{song2009joint}
Song, P. X.-K., Li, M., and Yuan, Y. (2009).
\newblock Joint regression analysis of correlated data using gaussian copulas.
\newblock {\em Biometrics}, 65(1):60--68.

\bibitem[Sun et~al., 2022]{sun2022multi}
Sun, K., He, M., Xu, Y., Wu, Q., He, Z., Li, W., Liu, H., and Pi, X. (2022).
\newblock Multi-label classification of fundus images with graph convolutional network and lightgbm.
\newblock {\em Computers in Biology and Medicine}, 149:105909.

\bibitem[Tideman et~al., 2016]{tideman2016association}
Tideman, J. W.~L., Snabel, M.~C., Tedja, M.~S., Van~Rijn, G.~A., Wong, K.~T., Kuijpers, R.~W., Vingerling, J.~R., Hofman, A., Buitendijk, G.~H., Keunen, J.~E., et~al. (2016).
\newblock Association of axial length with risk of uncorrectable visual impairment for europeans with myopia.
\newblock {\em JAMA ophthalmology}, 134(12):1355--1363.

\bibitem[Yang et~al., 2019]{yang2019nonparametric}
Yang, L., Frees, E.~W., and Zhang, Z. (2019).
\newblock Nonparametric estimation of copula regression models with discrete outcomes.
\newblock {\em Journal of the American Statistical Association}.

\bibitem[Yang et~al., 2022]{yang2022image}
Yang, S., Xiao, W., Zhang, M., Guo, S., Zhao, J., and Shen, F. (2022).
\newblock Image data augmentation for deep learning: A survey.
\newblock {\em arXiv preprint arXiv:2204.08610}.

\bibitem[Zellner, 1962]{zellner1962efficient}
Zellner, A. (1962).
\newblock An efficient method of estimating seemingly unrelated regressions and tests for aggregation bias.
\newblock {\em Journal of the American statistical Association}, 57(298):348--368.

\bibitem[Zhang et~al., 2024]{zhang2024axial}
Zhang, S., Chen, Y., Li, Z., Wang, W., Xuan, M., Zhang, J., Hu, Y., Chen, Y., Xiao, O., Yin, Q., et~al. (2024).
\newblock Axial elongation trajectories in chinese children and adults with high myopia.
\newblock {\em JAMA ophthalmology}, 142(2):87--94.

\bibitem[Zou et~al., 2022]{zou2022estimation}
Zou, C., Ke, Y., and Zhang, W. (2022).
\newblock Estimation of low rank high-dimensional multivariate linear models for multi-response data.
\newblock {\em Journal of the American Statistical Association}, 117(538):693--703.

\end{thebibliography}
\end{document}